\documentclass[a4paper,twoside]{article}

\usepackage{epsfig}
\usepackage{subcaption}
\usepackage{calc}
\usepackage{amssymb}
\usepackage{amstext}
\usepackage{amsmath}
\usepackage{amsthm}
\usepackage{multicol}
\usepackage{pslatex}
\usepackage{apalike}
\usepackage{algorithm2e}
\usepackage[bottom]{footmisc}

\usepackage{glossaries}
\usepackage{color,soul}
\usepackage{graphicx}

\newacronym{ros}{ROS}{Robot Operative System}
\newacronym{dft}{DFT}{Discrete Fourier Transform}
\newacronym{idft}{IDFT}{Inverse Discrete Fourier Transform}
\newacronym{iou}{IOU}{Intersection Over Union}
\newacronym{bb}{BB}{Bounding Box}
\newacronym{dof}{DoF}{Degrees of Freedom}
\newacronym[plural=NNs, firstplural=Neural Networks (NNs)]{nn}{NN}{Neural Network}
\newacronym{sam}{SAM}{Segment Anything Model}
\newacronym{msi}{MSI}{MultiSpectral Imaging}
\newacronym{lidar}{LiDAR}{Light Detection And Ranging}
\newacronym[plural=GPUs, firstplural=Graphics Processing Units (GPUs)]{gpu}{GPU}{Graphics Processing Unit}
\newacronym{ucm}{UCM}{UC Merced land use }
\newacronym{sift}{SIFT}{Scale-Invariant Feature Transform}
\newacronym{cpu}{CPU}{Central Processing Unit}

\usepackage{SCITEPRESS}     

\begin{document}

\title{Transferability of labels between multilens cameras}

\author{\authorname{Ignacio de Loyola Páez-Ubieta \sup{1}\orcidAuthor{0000-0001-9901-7264}, Daniel Frau-Alfaro\sup{1}\orcidAuthor{0009-0000-4098-3783} and Santiago T. Puente\sup{1}\orcidAuthor{0000-0002-6175-600X}}
\affiliation{\sup{1}AUtomatics, RObotics, and Artificial Vision (AUROVA) Lab, University Institute for Computer Research (IUII), University of Alicante, Crta. San Vicente s/n, San Vicente del Raspeig, E-03690, Alicante, Spain}
\email{\{ignacio.paez, daniel.frau, santiago.puente\}@ua.es}
}



\abstract{
In this work, a new method for automatically extending Bounding Box (BB) and mask labels across different channels on multilens cameras is presented. For that purpose, the proposed method combines the well known phase correlation method with a refinement process. During the first step, images are aligned by localizing the peak of intensity obtained in the spatial domain after performing the cross correlation process in the frequency domain. The second step consists of obtaining the best possible transformation by using an iterative process maximising the IoU (Intersection over Union) metric. Results show that, by using this method, labels could be transferred across different lens on a camera with an accuracy over 90\% in most cases and just by using 65 ms in the whole process. Once the transformations are obtained, artificial RGB images are generated, for labeling them so as to transfer this information into each of the other lens. This work will allow users to use this type of cameras in more fields rather than satellite or medical imagery, giving the chance of labeling even invisible objects in the visible spectrum.     
}

\onecolumn \maketitle \normalsize \setcounter{footnote}{0} \vfill

\section{\uppercase{Introduction}}
\label{sec:introduction}
When training a detection \cite{introyolov7} or segmentation \cite{introsolov2} \gls{nn}, a large amount of data is required to adapt an already trained model to a specific task, such as household waste \cite{introhouseholddetection} \cite{introhouseholdsegmentation}. 

However, times have changed and automatic labeling models for objects in RGB images appeared recently, with \gls{sam} model \cite{introsam} becoming a reference in a very short period of time.

\gls{msi} consists of using sensors that provide images in different frequency ranges compared to the traditional RGB cameras. In several fields such as in agriculture \cite{intromultiagricultureold} \cite{intromultiagriculturenew}, medical \cite{intromultimedicineold} \cite{intromultimedicinenew} or remote sensing \cite{intromultiremoteold} \cite{intromultiremotenew} these type of cameras have given promising results during the last two centuries. 

However, labeling images outside the RGB domain could be a difficult task. However, little by little, more and more articles dealing with the labeling process on \gls{lidar} \cite{introsamlidar} or multilens \cite{introsammulti} images are being published. 

For instance, \cite{compareClassificationLabels} introduced a semisupervised learning approach to automatically classify scenes from land datasets such as EuroSAT \cite{eurosat} or the aerial \gls{ucm} \cite{ucm}. For that purpose, they label between 5 and 300 images per class, which are then feed into a \gls{gpu} for training a model. In our case, we do not need any kind of training phase since we directly obtain the transformation between the camera lenses in order to have a more detailed object recognition rather than just a scene classification. Also, 15 images were used during the transformation phase, but less images could also be used for the proposed method.

Another example is \cite{compareSegmentateTrees}, in which authors aim to segmentate semantically trees using satellite and aerial images from the DSTL Satellite Imagery Feature Detection Image \cite{dstldataset} and RIT-18 (The Hamlin State Beach Park) Aerial Image \cite{rit18dataset} datasets. To this end, they use several segmentation \glspl{nn} to perform the task of labeling trees on the images. In our case, we do not require any kind of semantic segmentation \gls{nn} to label our images. Also, their trees occupy big areas on the images, making it easier for the \gls{nn} to find and label them, meanwhile the objects in our case are far smaller - tougher to label as they require far more precision and attention to fine detail.

Other works such as \cite{compareSIFT}, use a multispectral and RGB cameras for detecting sick pine trees through aerial photographs. For performing the image alignment for later labeling process they use the \gls{sift} method \cite{siftmethod}. However, this aforementioned method only works well when keypoints and descriptors could be obtained from the images, being useless with uniform objects. Also, RGB and 6 channel multispectral images are analysed by the \gls{nn}, making the process not very efficient since some of the 9 channels could contain no info at all.  

In this work, the transformation between the images captured from a multispectral camera (which is a type of multilens camera) will be obtained with the purpose of extending the \gls{bb} or mask labels from one image into the others, as well as allowing the users to label in fake RGB images, saving a significant amount of time. The process consists of a two step process (displacement calculation and refinement) that uses traditional computer vision techniques and \gls{cpu} resources, leaving aside time and resource consumption on \glspl{gpu}.

The main contributions of this work are:
\begin{itemize}
  \item A new method for obtaining the transformation between the lens of a multispectral camera that is proven to be highly accurate in no time. 
  \item The possibility of generating fake RGB images from combining its components by applying the previous transform. 
  \item Transforming labels in both \gls{bb} and mask formats across images so as to label objects that disappear in certain frequencies.  
\end{itemize}

This work is organised as follows: Section \ref{sec:methodology} introduces the proposed method, which is divided in two steps, Section \ref{sec:experimentation} presents the setup that was used during the experiments, as well as the transformations between each lenses and the fake RGB labeling process and Section \ref{sec:conclusion} summarises the article and introduces future works that will be done using as core project this method.

\section{\uppercase{Methodology}}
\label{sec:methodology}
In this Section, the method for obtaining the transformation between different lens on the camera is presented. It is composed of two steps: displacement calculation, using the phase correlation method, and refinement, using a sliding window across several scales.

\subsection{Displacement calculation}
\label{subsec:meth_dispCalc}
The different lens on a camera provide images that are not aligned. As the lenses are at the same height, a 2D transformation (rotation, translation, scale and/or skew) is the most probable conversion to relate them. However, the assumption that there is only a displacement between the captured images by the different lenses was done. If results proved otherwise, other kind of transformation would be applied.

The obtained images are in the space domain, in which each pixel represents the intensity. However, we switch into the frequency domain, in which images are reorganised by frequencies, distributing them according to its periodicity (high periodicity will be placed in the center of the image, meanwhile low periodicity will be placed far from it). 

By getting advance of how images are distributed in the frequency domain, the displacement between two images is a linear phase shift, which is the core idea of the phase correlation algorithm. 

It receives as input two images $i_1$ and $i_2$. The first step is removing sharp discontinuities at the image borders, since they produce a high frequency component, reducing the accuracy of the method. This problem is called spectral leakage, but applying a Hanning window (Eq. \ref{eq:hann}) will make it disappear, smoothing the image and removing artifacts and edges.

\begin{equation}
\label{eq:hann}
\begin{split}    
  w\left( x,y \right) = \left( 0.5 \left( 1 - cos\left( \frac{2 \pi x}{M - 1} \right) \right) \right) \cdot \\
  \left(0.5\left( 1-cos\left( \frac{2\pi y}{N-1} \right) \right)  \right)  
\end{split}
\end{equation}

where $M$ and $N$ represent the dimensions of the image and $x$ and $y$ represent the pixel coordinates.  
If applied to the aforementioned images, $i_{1h} (x,y)$ and $i_{2h} (x,y)$ are obtained. The second step consists of in transforming these spectral leakage free images into the frequency domain by using the \gls{dft} (Eq. \ref{eq:dft}), obtaining $I_{1h} (u,v)$ and $I_{2h} (u,v)$, respectively.

\begin{equation}
\label{eq:dft}
   I_{1h} (u,v) = \sum_{x=0}^{M-1}\sum_{y=0}^{N-1} i_{1h} (x,y) \cdot e^{-2\pi i \left( \frac{ux}{M} + \frac{vy}{N} \right)} 
\end{equation}

Once the images are in the frequency domain, the phase shift between them encode the translational shift we are looking for in the space domain. For that purpose, the third step is to isolate this phase information by using the cross-power spectrum (Eq. \ref{eq:cross_power}), normalising the magnitude and retaining this phase information. 

\begin{equation}
\label{eq:cross_power}
   CP\left( u,v \right) = \frac{I_{1h} (u,v) \cdot I_{2h}^* (u,v)}{\left| I_{1h} (u,v) \cdot I_{2h}^* (u,v) \right|}
\end{equation}

where $I_{2h}^* (u,v)$ is the complex conjugate of $I_{2h} (u,v)$. The forth step is coming back to the spatial domain by using the \gls{idft} applied to the calculated cross-power spectrum $CP\left( u,v \right)$, obtaining the correlation matrix $c \left( x,y \right)$ (Eq. \ref{eq:ift}).

\begin{equation}
\label{eq:ift}
   c (x,y) = \sum_{u=0}^{M-1}\sum_{v=0}^{N-1} CP \left(u,v \right) \cdot e^{2\pi i \left( \frac{ux}{M} + \frac{vy}{N} \right)} 
\end{equation}

Finally, peak location $(\Delta x, \Delta y)$ in the correlation matrix  $c \left( x,y \right)$ is obtained  (Eq. \ref{eq:peakLocation}) by performing a $5 \times 5$ weighted centroid around the peak, so as to obtain subpixel accuracy, normalizing the result between 0 and 1.

\begin{equation}
\label{eq:peakLocation}
    (\Delta x, \Delta y) = weightedCentroid \{\arg \max_{(x, y)}\{ c \left( x,y \right)\}\} 
\end{equation}

\subsection{Refinement}
\label{subsec:meth_refinement}
Once the relative displacement $(\Delta x, \Delta y)$ between the two input images $i_1, i_2$ is obtained using the phase correlation method, a refinement process is required in order to get more precision. 

For that purpose, centered in the relative displacement numbers, a cascade of possible better displacements are obtained. First, $(\Delta x, \Delta y)$ are rounded and are added or substracted a value $RV : i \in 1 \dots n$ with $n$ being the number of refinement steps in both axis $x$ and $y$ in different scales $s$. This $s$ value will represent different orders of magnitude, varying it inside the discrete values on $[1, 0.1, 0.01]$ to check pixel and subpixel precision (Eq. \ref{eq:s_delta}), having as a result several possible combinations. 

\begin{equation}
\label{eq:s_delta}
\begin{split}
    \Delta{x_p} = \left[ \Delta{x}-RV \cdot s, \cdots, \Delta{x}, \cdots \Delta{x}+RV \cdot s \right] \\
    \Delta{y_p} = \left[ \Delta{y}-RV \cdot s, \cdots, \Delta{y}, \cdots \Delta{y}+RV \cdot s \right]
\end{split}
\end{equation}

These obtained possible numbers $(\Delta{x_p}, \Delta{y_p})$ are inserted into a homogeneous transformation (Eq. \ref{eq:hom}) and applied to each of the different labels on the image to check if a better solution is obtained. For that purpose, some labeled images serve as basis to compare with these newly obtained labels. 

\begin{equation}
\label{eq:hom}
 \begin{bmatrix}
   l_{N:nx} \\
   l_{N:ny} 
 \end{bmatrix}
 = 
 \begin{bmatrix}
   1 & 0 & \Delta{x_p}  \\
   0 & 1 & \Delta{y_p} \\
 \end{bmatrix} 
 \begin{bmatrix}
   l_{nx} \\
   l_{ny}
  \end{bmatrix} 
\end{equation}

These labels $l = [(l_{1x},l_{1y}), ... , (l_{nx},l_{ny}), ..., (l_{Nx},l_{Ny})]$ are a collection of $N$ points that represent the borders of the labeled objects. In case of having a mask, $N$ could be any positive number, meanwhile a \gls{bb} is defined by $N$ = 2, representing the top left and bottom right coordinates of the box.

The final transform will convert the original mask or \gls{bb} coordinates $(l_{nx}, l_{ny})$ into the new reference frame, obtaining $l_{M} = [(l_{M:1x},l_{M:1y}), ... , (l_{M:nx},l_{M:ny}), ..., (l_{M:Nx},l_{M:Ny})]$. They will be compared against the aforementioned labeled images $l_{GT} = [(l_{GT:1x},l_{GT:1y}), ... , (l_{GT:nx},l_{GT:ny}), ..., (l_{GT:Nx},l_{GT:Ny})]$, trying to achieve the highest \gls{iou}.

\gls{iou}, also called Jaccard index, is a metric that returns how much two labels coincide, being the result a value between [0,1]. It is represented by Eq. \ref{eq:iou}.

\begin{equation}
\label{eq:iou}
IoU = \frac{l_M \cap l_{GT}}{l_{M} \cup l_{GT}}
\end{equation}

\section{\uppercase{Experimentation}}
\label{sec:experimentation}
In this Section, hardware and software setup is presented, as well as the experiments that prove the effectiveness of the proposed method for obtaining labels across multiple multispectral images.

\subsection{Setup}
\label{subsec:res_setup}
In order to perform image alignment, several instances are required. 

\begin{figure}[h]

    \begin{subfigure}[b]{0.5\textwidth}
        \centering
        \includegraphics[width=\textwidth]{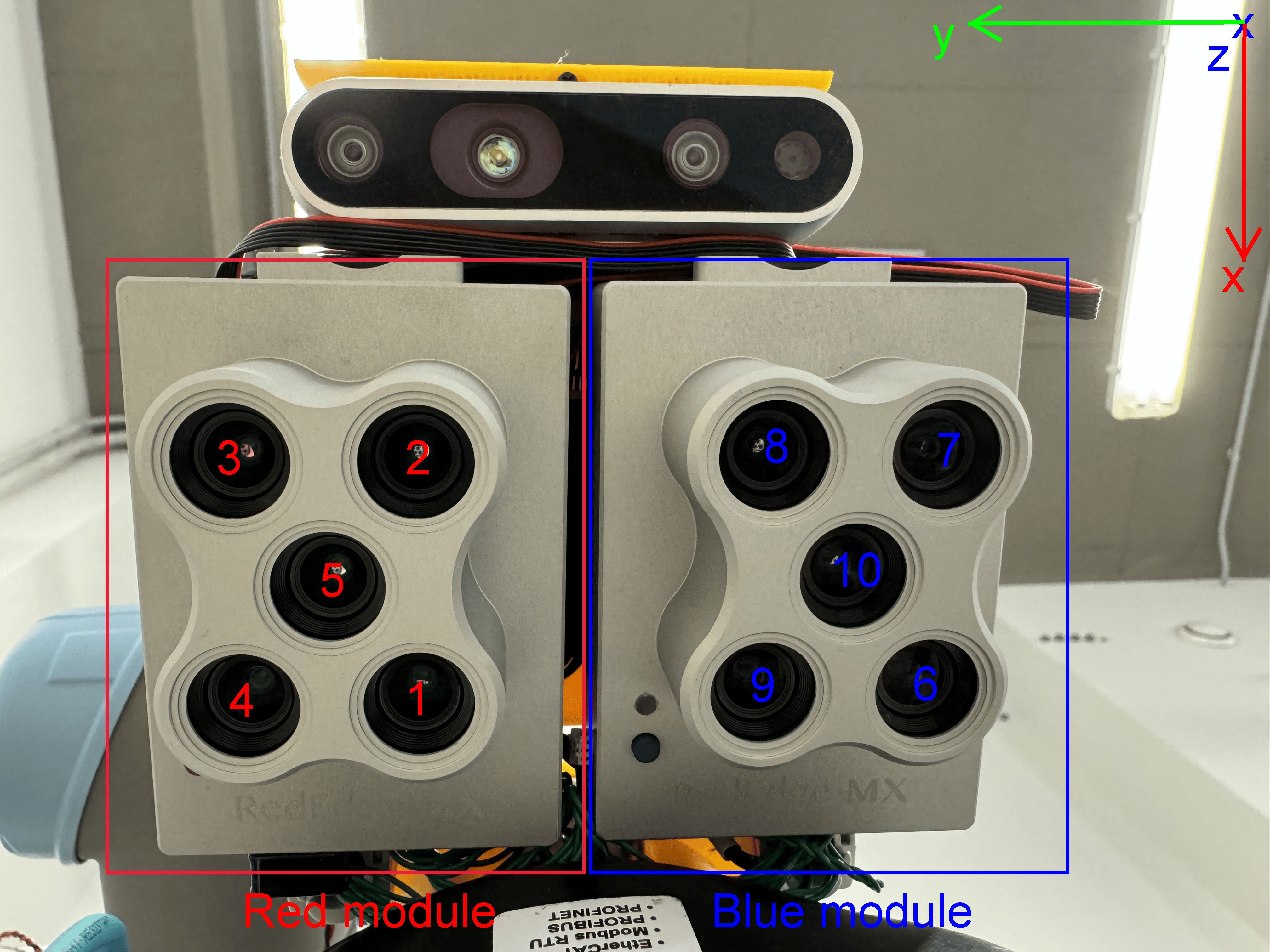}
        \caption{Detail of the multispectral camera lenses (modules and order).}
        \label{fig:SetupA}
    \end{subfigure}

    \hspace{0.1cm}

    \begin{subfigure}[b]{0.5\textwidth}
        \centering
        \includegraphics[trim={0cm 0cm 0cm 40cm}, clip, width=\textwidth]{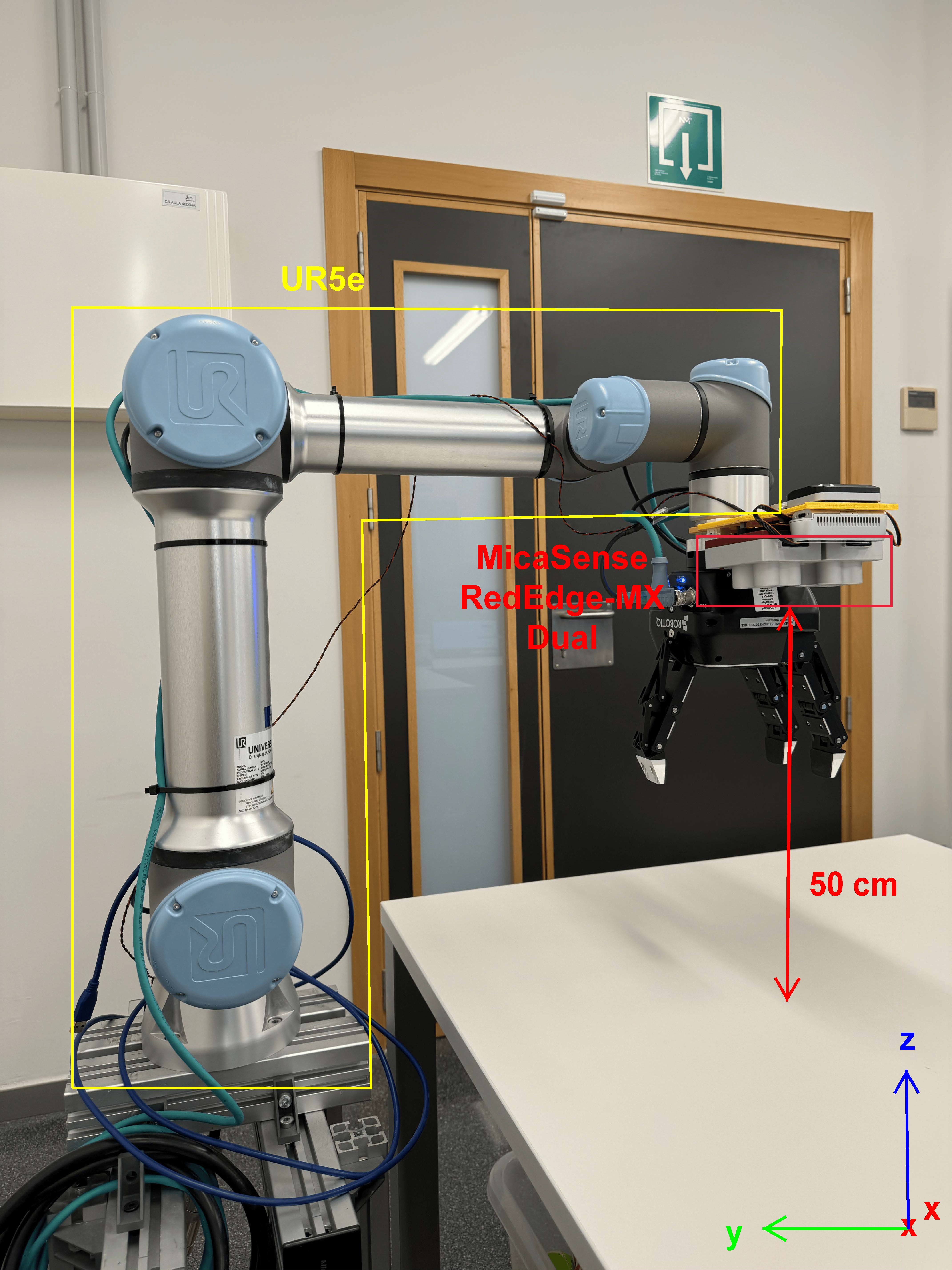}
        \caption{Robotic arm in which the MicaSense RedEdge-MX Dual is attached to, allowing to capture objects 50 cm away.}
        \label{fig:setupB}
    \end{subfigure}
    
    \hspace{1cm}
    
    \caption{Hardware used during the experiments.} \label{fig:setup}
\end{figure}

In terms of hardware (Fig. \ref{fig:setup}), a MicaSense RedEdge-MX Dual multispectral camera is used for obtaining the aforementioned images. This camera counts with 10 cameras divided in two modules: a red and a blue one (see Fig. \ref{fig:SetupA}). A brief description of the frequencies for each band available in both of them is presented in Table \ref{tab:wavelengths}. All 10 bands produce 12 bit images at a resolution of 1280 x 960. This camera is mounted on the wrist of an Ur5e with 6 \gls{dof} robotic arm, which allow us to precisely place the camera in a still position on the space. Concretely, the camera is positioned parallel to a table 500 mm over it (see Fig. \ref{fig:setupB}). Regarding the objects used for performing the experiments, 16 small pills in the size range of 8 to 22 mm are used. They provide several different shapes and colors to allow all the experiments that will be performed. 

\begin{table}[h]
\centering
\caption{Band numbers, frequencies and color name from each channel available on the MicaSense RedEdge-MX Dual multispectral camera (see Fig. \ref{fig:SetupA}).}
\label{tab:wavelengths}
\begin{tabular}{|c|c|c|c|}
\cline{1-4}
Module                    & Band & f ± A (nm)    & Color name   \\ \hline
\multicolumn{1}{|c|}{Red} & 1    & 475 ± 16      & Blue         \\ \cline{2-4} 
\multicolumn{1}{|c|}{   } & 2    & 560 ± 13.5    & Green        \\ \cline{2-4} 
\multicolumn{1}{|c|}{   } & 3    & 668 ± 7       & Red          \\ \cline{2-4} 
\multicolumn{1}{|c|}{   } & 4    & 717 ± 6       & Red Edge     \\ \cline{2-4} 
\multicolumn{1}{|c|}{   } & 5    & 842 ± 28.5    & Near IR      \\ \hline
\multicolumn{1}{|c|}{Blue}& 6    & 444 ± 14      & Coastal Blue \\ \cline{2-4} 
\multicolumn{1}{|c|}{    }& 7    & 531 ± 7       & Green        \\ \cline{2-4} 
\multicolumn{1}{|c|}{    }& 8    & 650 ± 8       & Red          \\ \cline{2-4} 
\multicolumn{1}{|c|}{    }& 9    & 705 ± 5       & Red Edge I   \\ \cline{2-4} 
\multicolumn{1}{|c|}{    }& 10   & 740 ± 9       & Red Edge II  \\ \hline
\end{tabular}
\end{table}

In terms of software, images were labeled using LabelMe tool \cite{labelme}. There are two different labeling approximations used across all the experiments: \gls{bb} and mask. The computer used to get the results works in Ubuntu 20.04.4 with Python 3.8.10 and OpenCV 4.7.0, running an 11th Generation Intel$^\copyright$ Core\texttrademark \hspace{0.1cm} i9-11900H with 8 physical and 16 logical cores respectively. They work at a frequency of 2.50 GHz, which allow to carry out all necessary operations in a short period of time.

\subsection{Transformations}
\label{subsec:res_transformations}
Although the camera has 10 lenses, only the ones in the red half part of the camera will be used. 

Several images from the 5 lenses that compose the red part of the multispectral camera are obtained. Concretely, 15 images were captured with each camera, from which 12 were used for obtaining the transform and 3 for proving the accuracy of the obtained transform. From them, the training images from band 5 (lenses in the middle) and the test images from all 5 lenses were labeled. Also, the refinement steps $n$ was set to 5, meaning 121 possible matrices in 3 different levels of pixel accuracy are used.

After applying the phase correlation method and the refinement for both \gls{bb} and mask labels, the following results shown in Table \ref{tab:bbTransform} and Table \ref{tab:maskTransform} were obtained. 

\begin{table}[hb]
\centering
\caption{Transformations in pixel level, \gls{iou} and time for \gls{bb} labels to transfer band 5 labels into the other lenses.}
\label{tab:bbTransform}
\begin{tabular}{|c|c|c|c|}
\hline
Band & Transform (px)                                 & IoU (\%) & Time (ms) \\ \hline
1    & $^1_5T_{BB} = \begin{pmatrix} -52.0 \\  47.0 \end{pmatrix}$ & 98.58    & 66.31     \\ \hline
2    & $^2_5T_{BB} = \begin{pmatrix}  53.9 \\  46.1 \end{pmatrix}$ & 100      & 53.52     \\ \hline
3    & $^3_5T_{BB} = \begin{pmatrix}  52.9 \\ -23.4 \end{pmatrix}$ & 95.95    & 55.66     \\ \hline
4    & $^4_5T_{BB} = \begin{pmatrix} -52.1 \\ -18.9 \end{pmatrix}$ & 93.53    & 56.55     \\ \hline
\end{tabular}
\end{table}

\begin{table}[ht]
\centering
\caption{Transformations in pixel level, \gls{iou} and time for mask labels to transfer band 5 labels into the other lenses.}
\label{tab:maskTransform}
\begin{tabular}{|c|c|c|c|}
\hline
Band & Transform (px)                                   & IoU (\%) & Time (ms) \\ \hline
1    & $^1_5T_{MK} = \begin{pmatrix} -52.05 \\  47.2  \end{pmatrix}$ & 97.49    & 82.81     \\ \hline
2    & $^2_5T_{MK} = \begin{pmatrix}  54.78 \\  46.52 \end{pmatrix}$ & 94.36    & 72.58     \\ \hline
3    & $^3_5T_{MK} = \begin{pmatrix}  53.5  \\ -23.8  \end{pmatrix}$ & 93.77    & 73.05     \\ \hline
4    & $^4_5T_{MK} = \begin{pmatrix} -53.24 \\ -19.01 \end{pmatrix}$ & 89.91    & 58.83     \\ \hline
\end{tabular}
\end{table}

In order to show the active refinement process and how it improves the \gls{iou} step by step, band 1 from mask labeling is shown in Table \ref{tab:refinement}. Step 0 consists of applying phase correlation, step 1 consists of refining in pixel level with $s$ equal to 1, step 2 consists of refining in subpixel level with $s$ equal to 0.1 and step 3 consists of refining in 2 levels of subpixel with $s$ equal to 0.01.

\begin{table}[h]
\centering
\caption{Phase correlation and refinement steps applied to band 1 of mask labeled images.}
\label{tab:refinement}
\begin{tabular}{|c|c|c|c|}
\hline
Step & Transform (px)                                   & IoU (\%) & Time (ms) \\ \hline
0    & $^1_5T_{MK-0} = \begin{pmatrix} -51.85 \\  47.02 \end{pmatrix}$ & 96.73    & 76.75     \\ \hline
1    & $^1_5T_{MK-1} = \begin{pmatrix} -52.0  \\  47.0  \end{pmatrix}$ & 96.99    &  1.96     \\ \hline
2    & $^1_5T_{MK-2} = \begin{pmatrix} -52.0  \\  47.2  \end{pmatrix}$ & 97.40    &  2.05     \\ \hline
3    & $^1_5T_{MK} = \begin{pmatrix} -52.05 \\  47.2  \end{pmatrix}$ & 97.49    &  2.05     \\ \hline
\end{tabular}
\end{table}

\begin{figure*}[ht]
\centering
     \begin{subfigure}[b]{0.24\textwidth}
         \centering
         \includegraphics[trim={24cm 15cm 10cm 12cm}, clip, width=\textwidth]{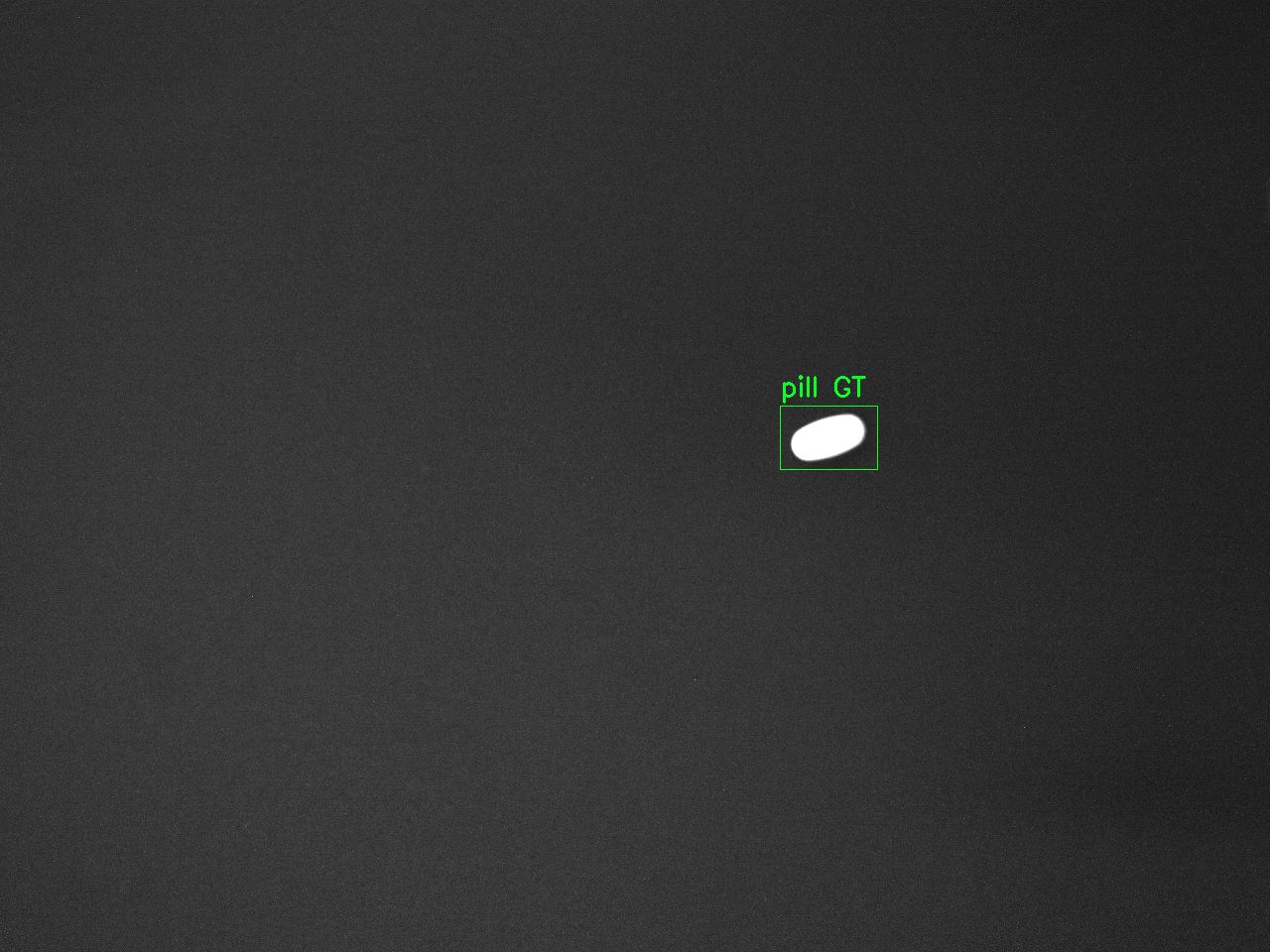}
         \caption{Band 5 input.}
         \label{fig:bb_input}
     \end{subfigure}
     
     \hspace{1cm}
     
     \centering
     \begin{subfigure}[b]{0.24\textwidth}
         \centering
         \includegraphics[trim={24cm 15cm 10cm 12cm}, clip, width=\textwidth]{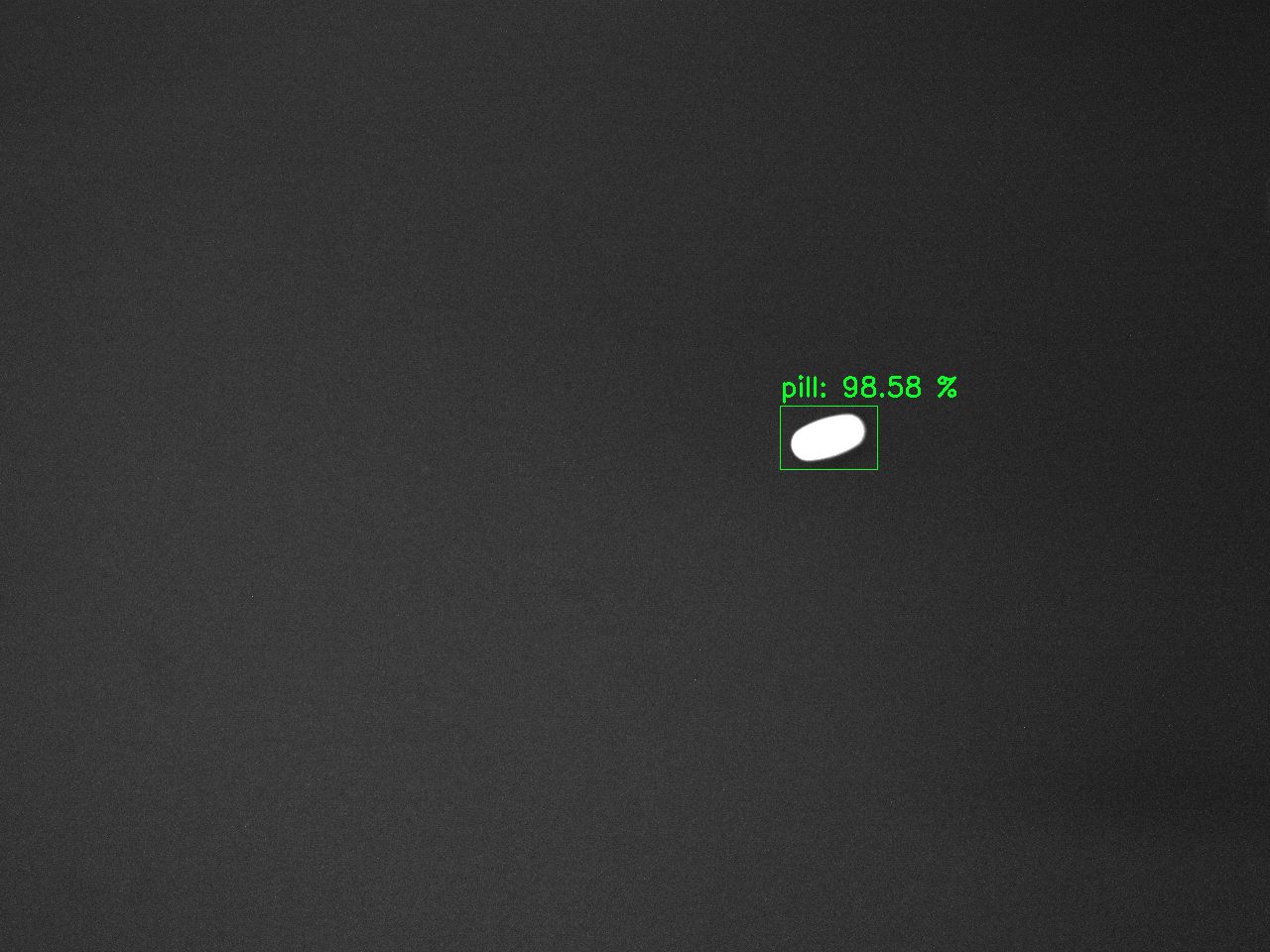}
         \caption{Band 1 output.}
         \label{fig:bb_out1}
     \end{subfigure}
     \begin{subfigure}[b]{0.24\textwidth}
         \centering
         \includegraphics[trim={28cm 15cm 6cm 12cm}, clip, width=\textwidth]{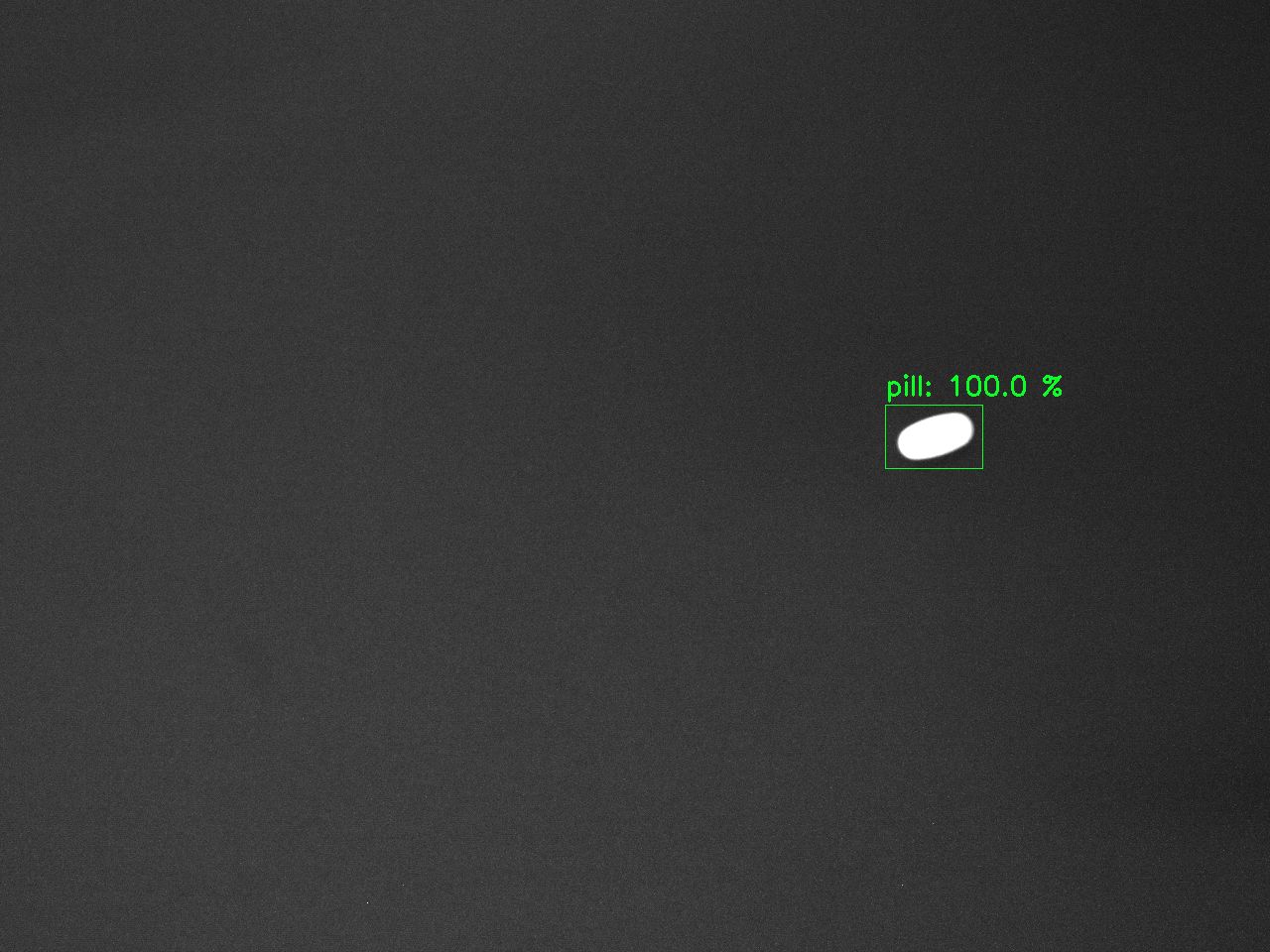}
         \caption{Band 2 output.}
         \label{fig:bb_out2}
     \end{subfigure}
     \begin{subfigure}[b]{0.24\textwidth}
         \centering
         \includegraphics[trim={27.5cm 17.5cm 6.5cm 9.5cm}, clip, width=\textwidth]{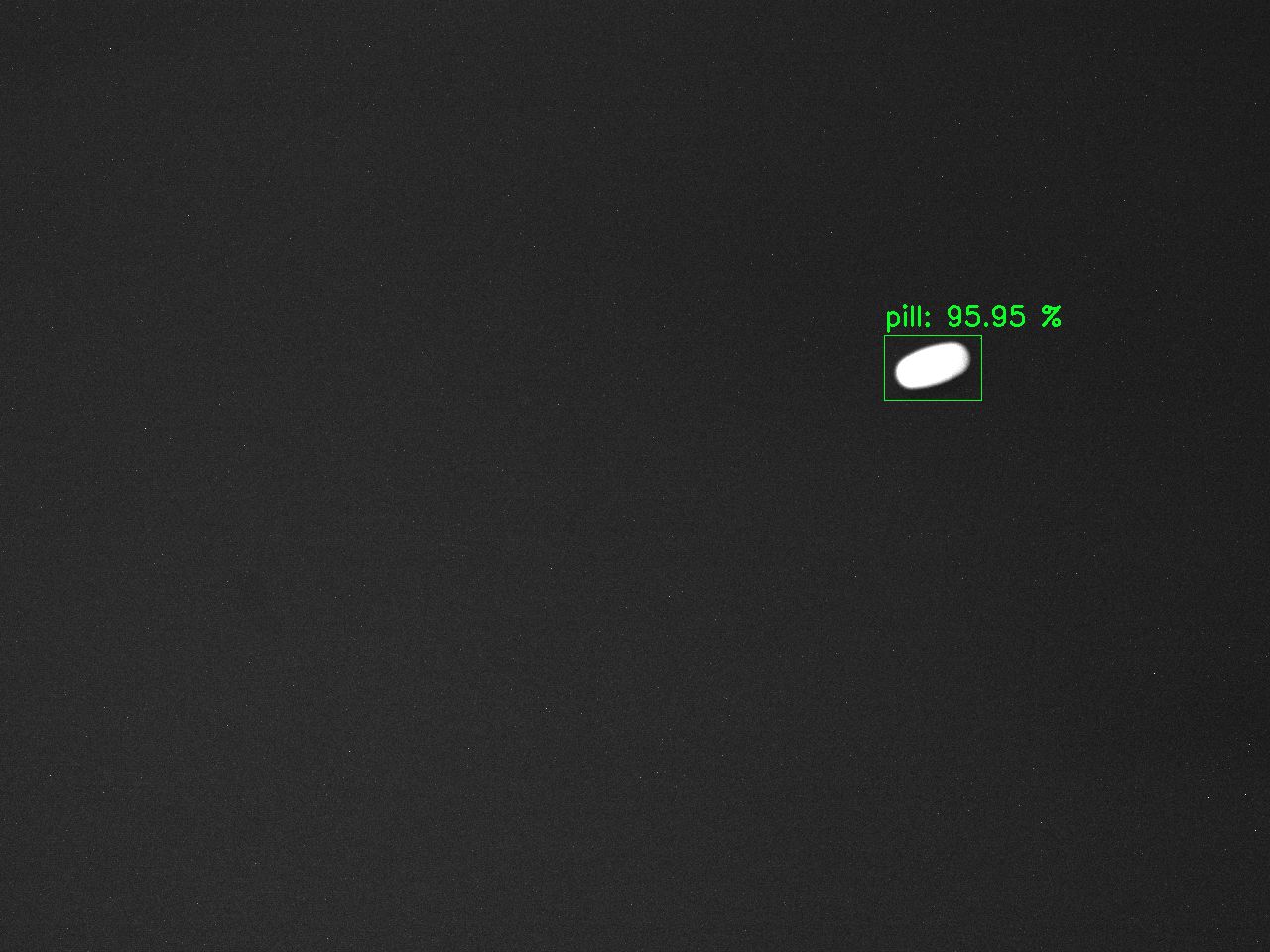}
         \caption{Band 3 output.}
         \label{fig:bb_out3}
     \end{subfigure}
     \begin{subfigure}[b]{0.24\textwidth}
         \centering
         \includegraphics[trim={23.5cm 17.5cm 10.5cm 9.5cm}, clip, width=\textwidth]{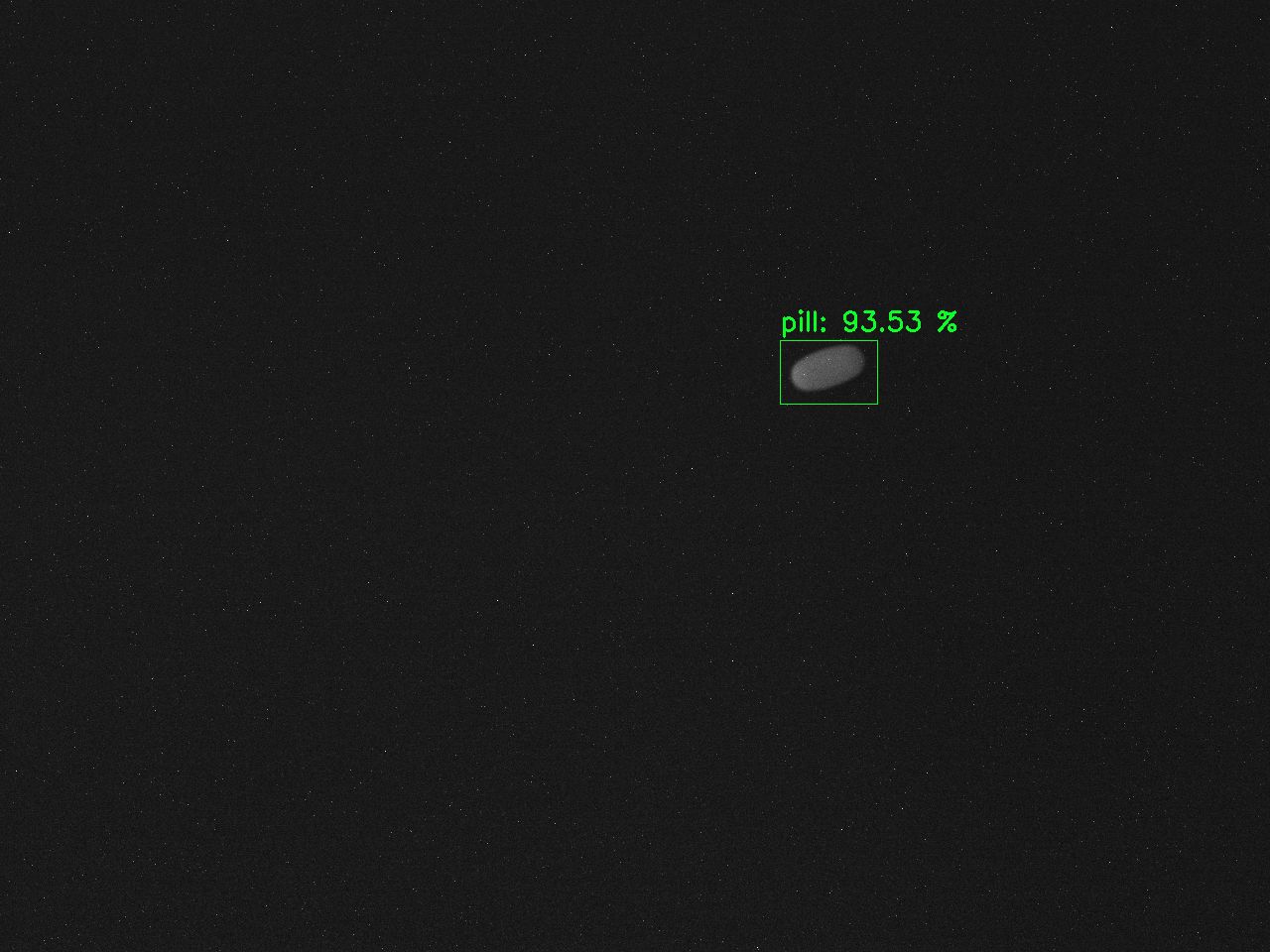}
         \caption{Band 4 output.}
         \label{fig:bb_out4}
     \end{subfigure}
     
     \hspace{1cm}
     
     \begin{subfigure}[b]{0.24\textwidth}
         \centering
         \includegraphics[trim={24cm 15cm 10cm 12cm},clip, width=\textwidth]{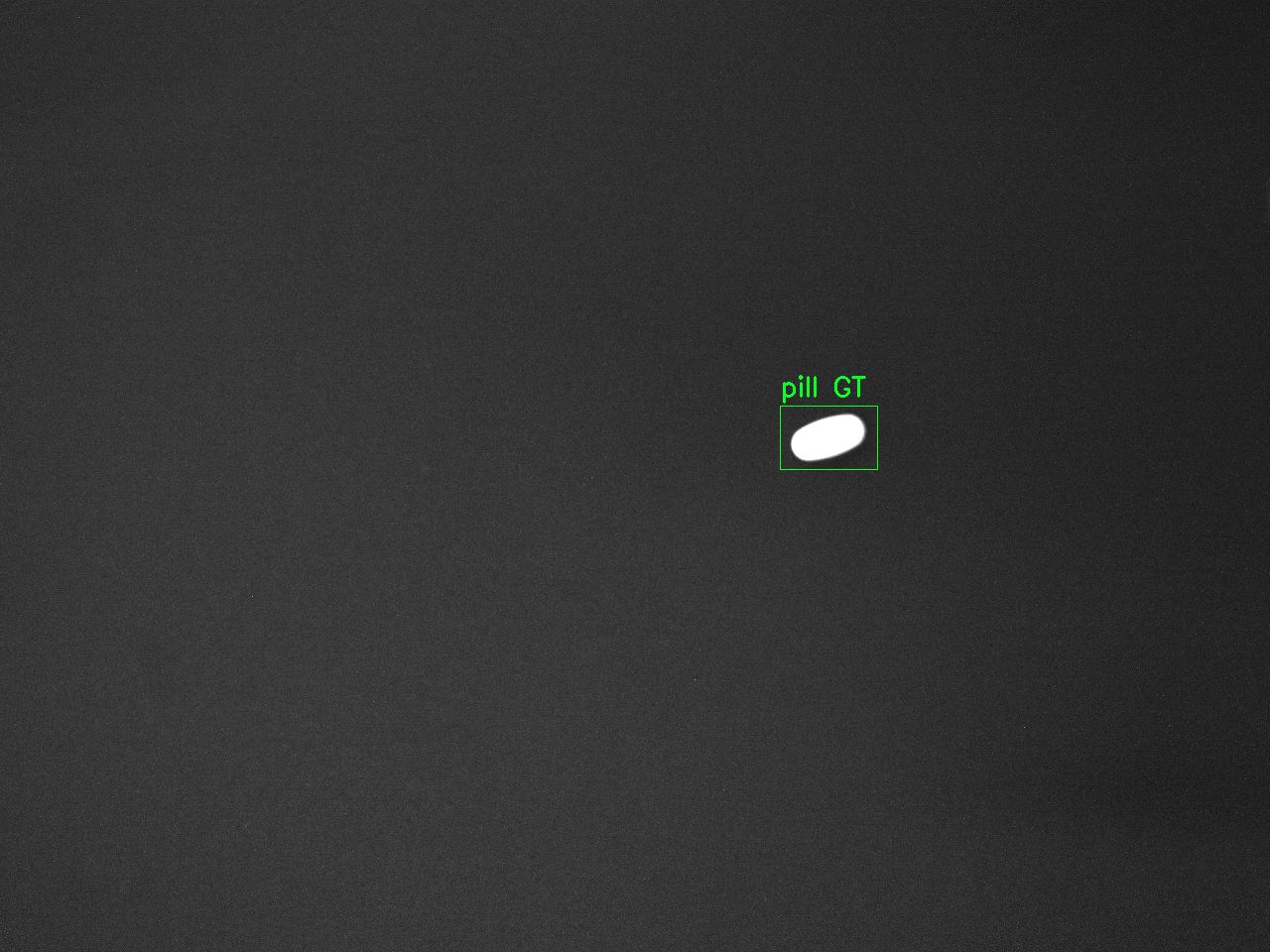}
         \caption{Band 1 reference.}
         \label{fig:bb_GT1}
     \end{subfigure}
     \begin{subfigure}[b]{0.24\textwidth}
         \centering
         \includegraphics[trim={28cm 15cm 6cm 12cm}, clip, width=\textwidth]{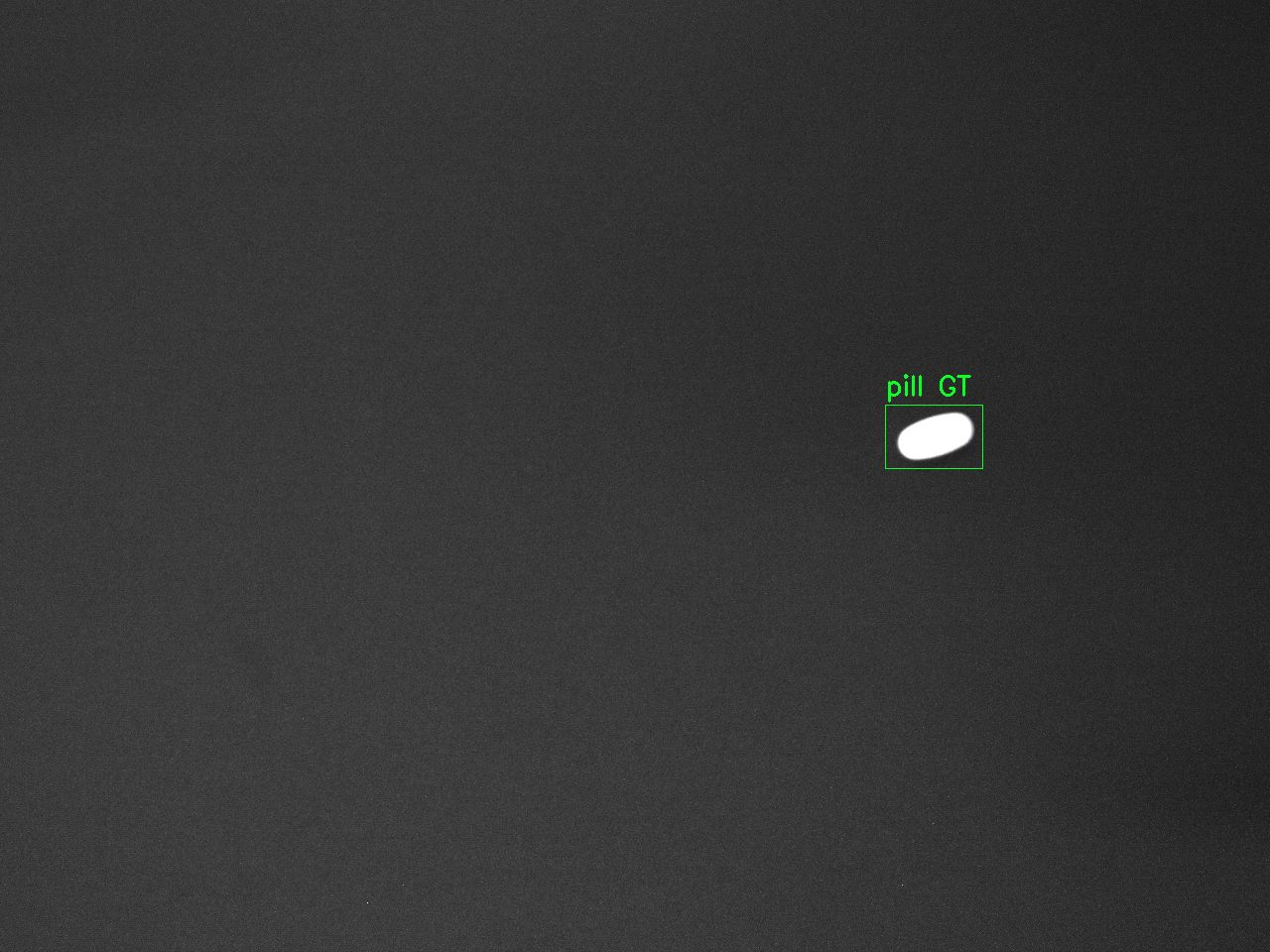}
         \caption{Band 2 reference.}
         \label{fig:bb_GT2}
     \end{subfigure}
     \begin{subfigure}[b]{0.24\textwidth}
         \centering
         \includegraphics[trim={27.5cm 17.5cm 6.5cm 9.5cm}, clip, width=\textwidth]{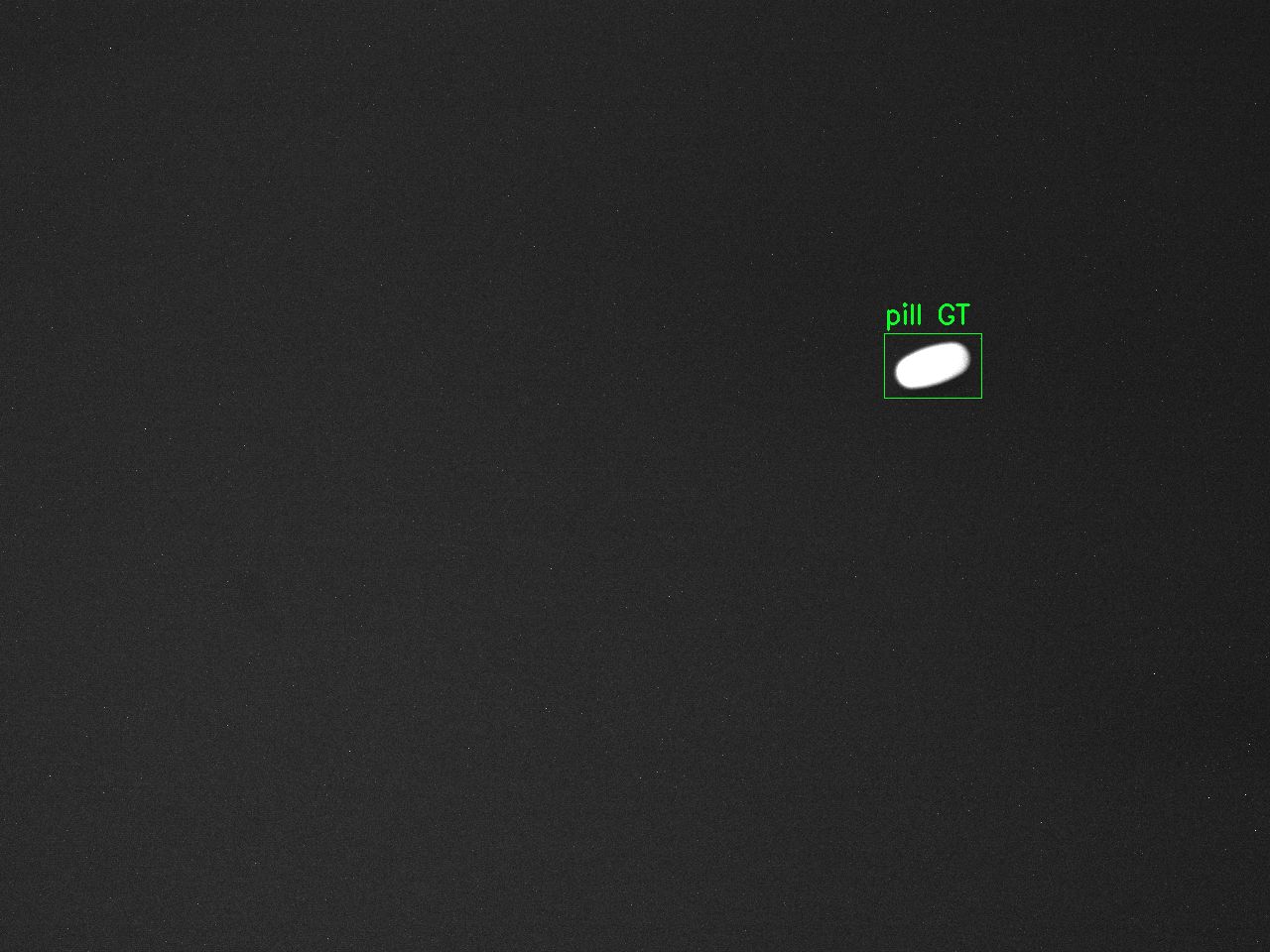}
         \caption{Band 3 reference.}
         \label{fig:bb_GT3}         
     \end{subfigure}
     \begin{subfigure}[b]{0.24\textwidth}
         \centering
         \includegraphics[trim={23.5cm 17.5cm 10.5cm 9.5cm}, clip, width=\textwidth]{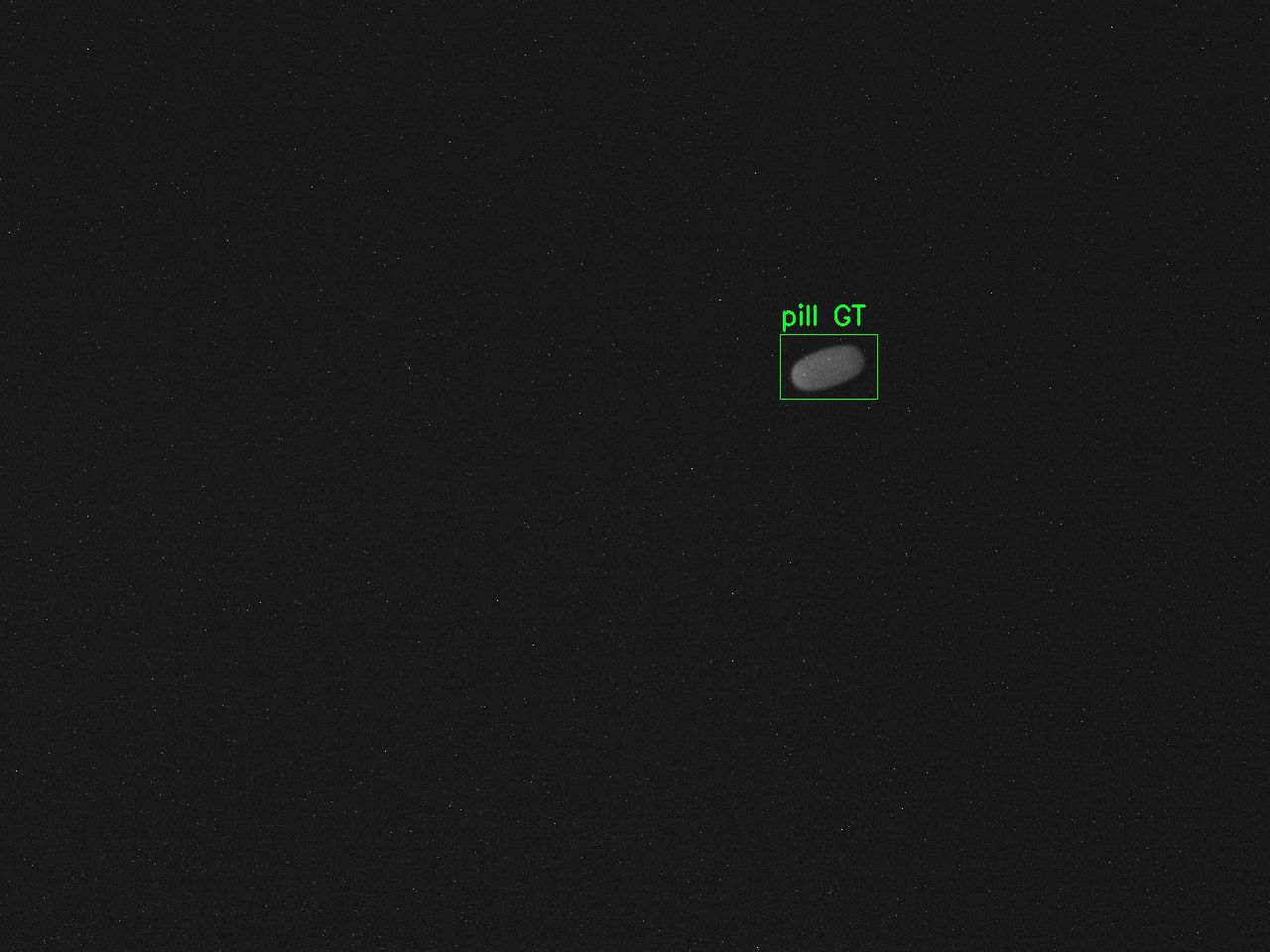}
         \caption{Band 4 reference.}
         \label{fig:bb_GT4}
     \end{subfigure}

     \hspace{1cm}
     
     \caption{\gls{bb} labeled experiment: (a) reference image (band 5) to start with, (b, c, d, e) transformed labels (bands 1-4) and (f, g, h, i) ground truth labels for comparison purposes (bands 1-4).}\label{fig:transfResBB}
\end{figure*}

Several images of \gls{bb} labeled images after applying $^1_5T_{BB}$, $^2_5T_{BB}$, $^3_5T_{BB}$ and $^4_5T_{BB}$ from Table \ref{tab:bbTransform} can be seen in Fig. \ref{fig:transfResBB}. As input Fig. \ref{fig:bb_input} is provided, which represents band 5. Once the labels of this image are transformed, images in Figs. \ref{fig:bb_out1}, \ref{fig:bb_out2}, \ref{fig:bb_out3} and \ref{fig:bb_out4} are obtained. As it can be seen, the transformed labels adjust almost perfectly into the objects of the other bands, saving the user the need of manually annotate them. For a quick comparison of the quality, ground truth human labeled images are provided in Figs. \ref{fig:bb_GT1}, \ref{fig:bb_GT2}, \ref{fig:bb_GT3} and \ref{fig:bb_GT4}. The worst result is obtained in band 4, since the pill starts mimicking with the background, making it difficult for both the proposed method and the user to discern it.

\begin{figure*}[!h]
\centering
     \begin{subfigure}[b]{0.24\textwidth}
         \centering
         \includegraphics[trim={28cm 16.5cm 6cm 10.5cm}, clip, width=\textwidth]{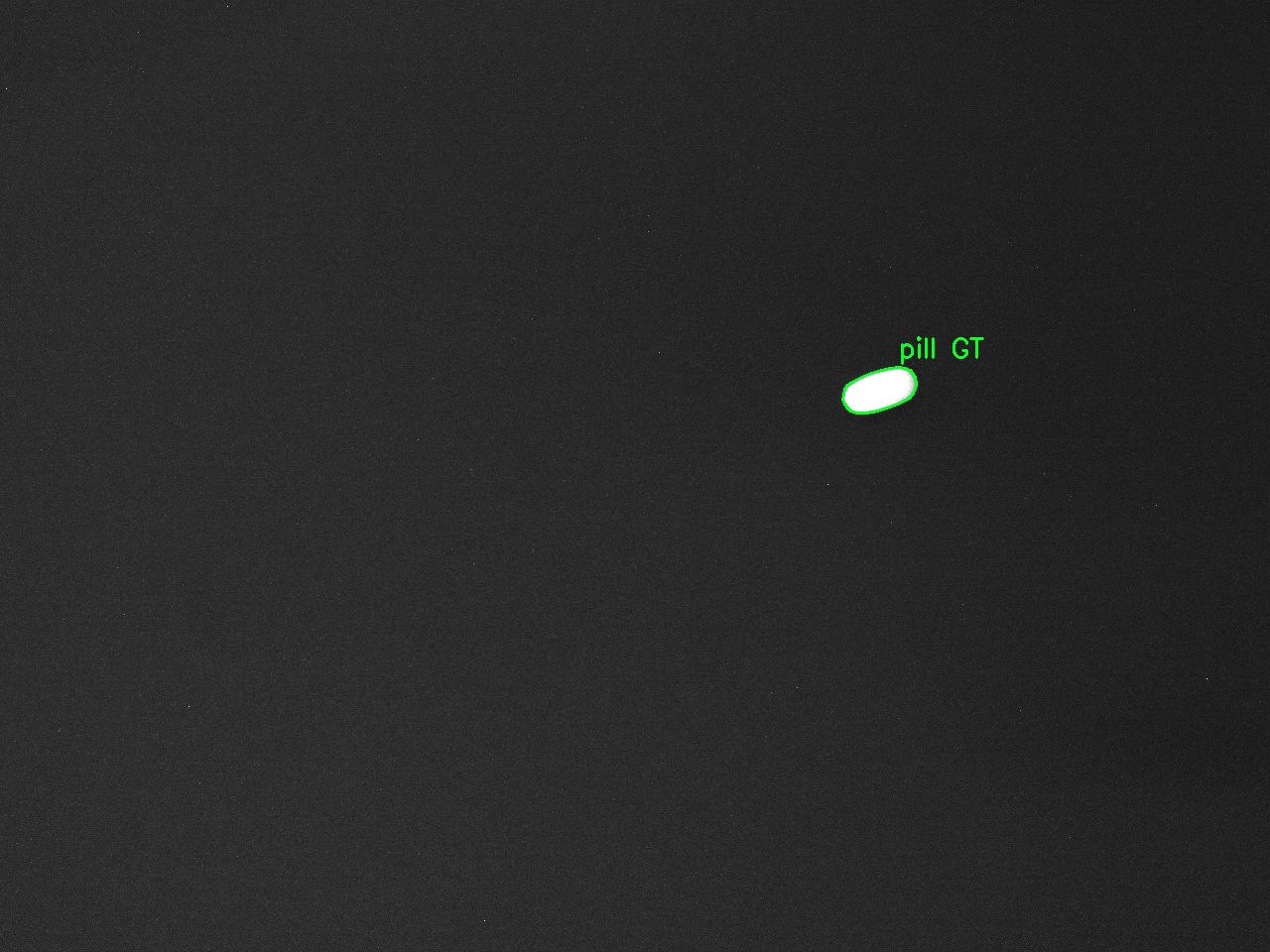}
         \caption{Band 5 input.}
         \label{fig:mask_input}
     \end{subfigure}
     
     \hspace{1cm}
     
     \centering
     \begin{subfigure}[b]{0.24\textwidth}
         \centering
         \includegraphics[trim={26cm 15cm 8cm 12cm}, clip, width=\textwidth]{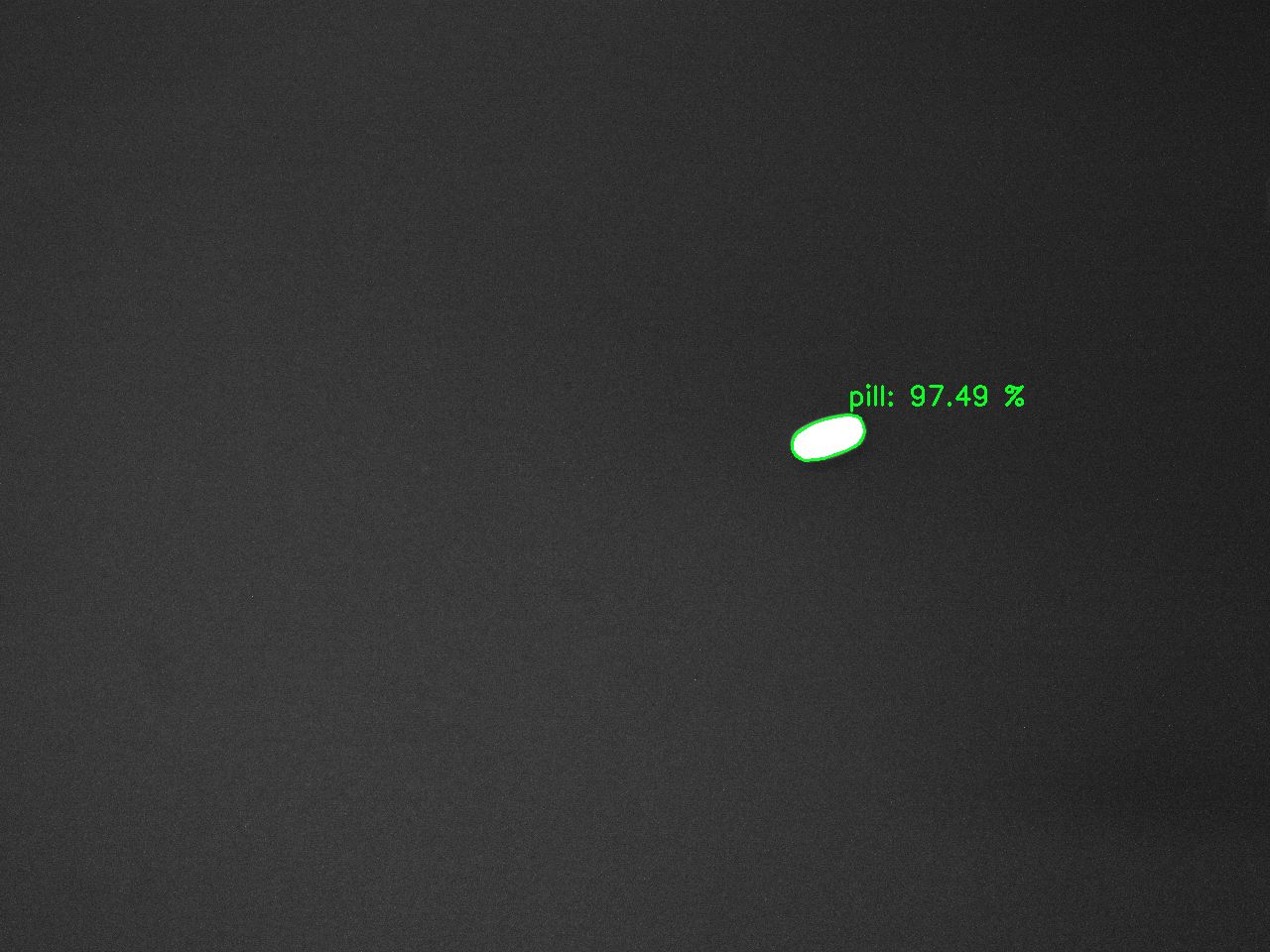}
         \caption{Band 1 output.}
         \label{fig:mask_out1}
     \end{subfigure}
     \begin{subfigure}[b]{0.24\textwidth}
         \centering
         \includegraphics[trim={30cm 15cm 4cm 12cm}, clip, width=\textwidth]{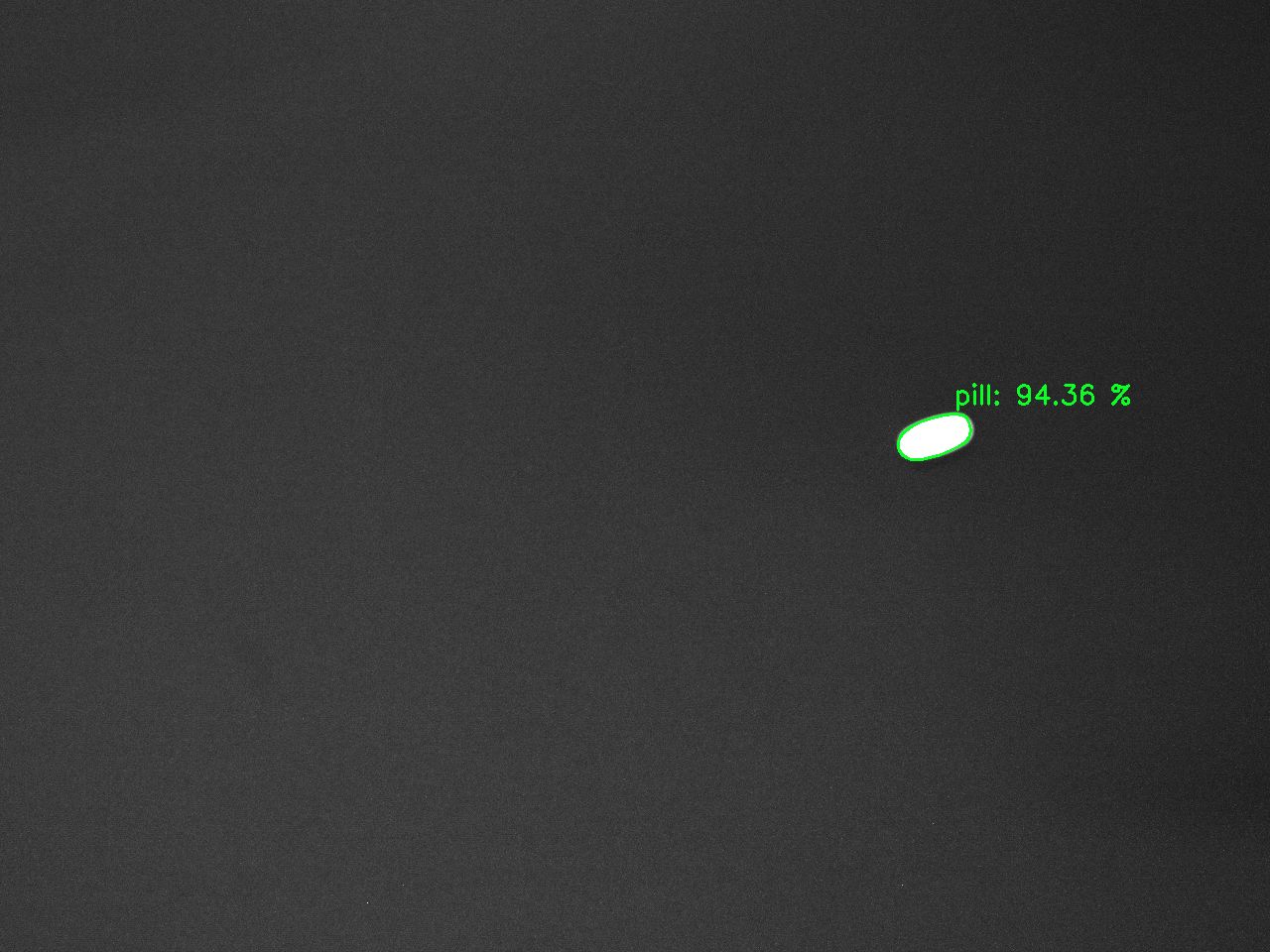}
         \caption{Band 2 output.}
         \label{fig:mask_out2}
     \end{subfigure}
     \begin{subfigure}[b]{0.24\textwidth}
         \centering
         \includegraphics[trim={29.5cm 17.5cm 4.5cm 9.5cm}, clip, width=\textwidth]{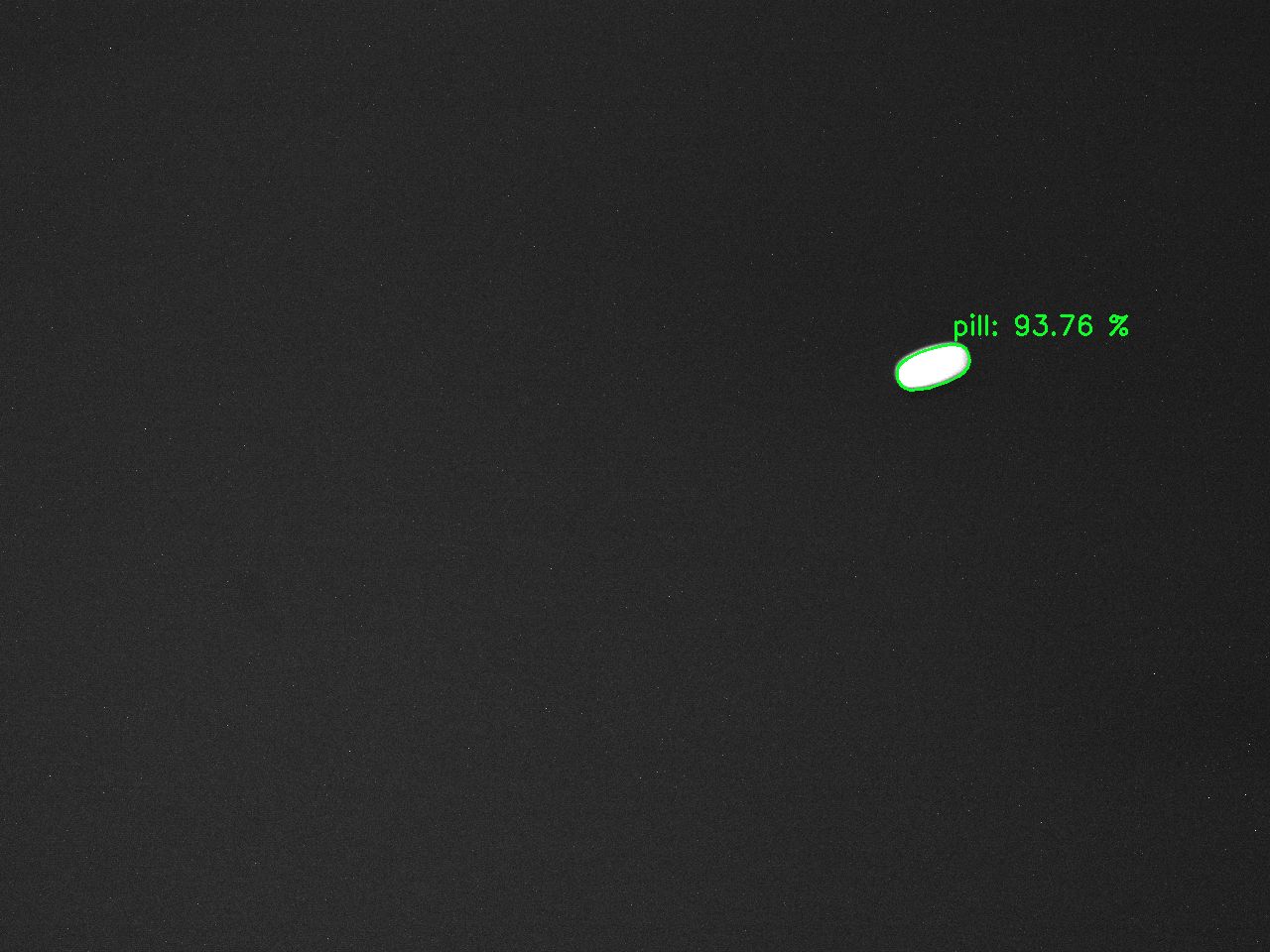}
         \caption{Band 3 output.}
         \label{fig:mask_out3}
     \end{subfigure}
     \begin{subfigure}[b]{0.24\textwidth}
         \centering
         \includegraphics[trim={25.5cm 17.5cm 8.5cm 9.5cm}, clip, width=\textwidth]{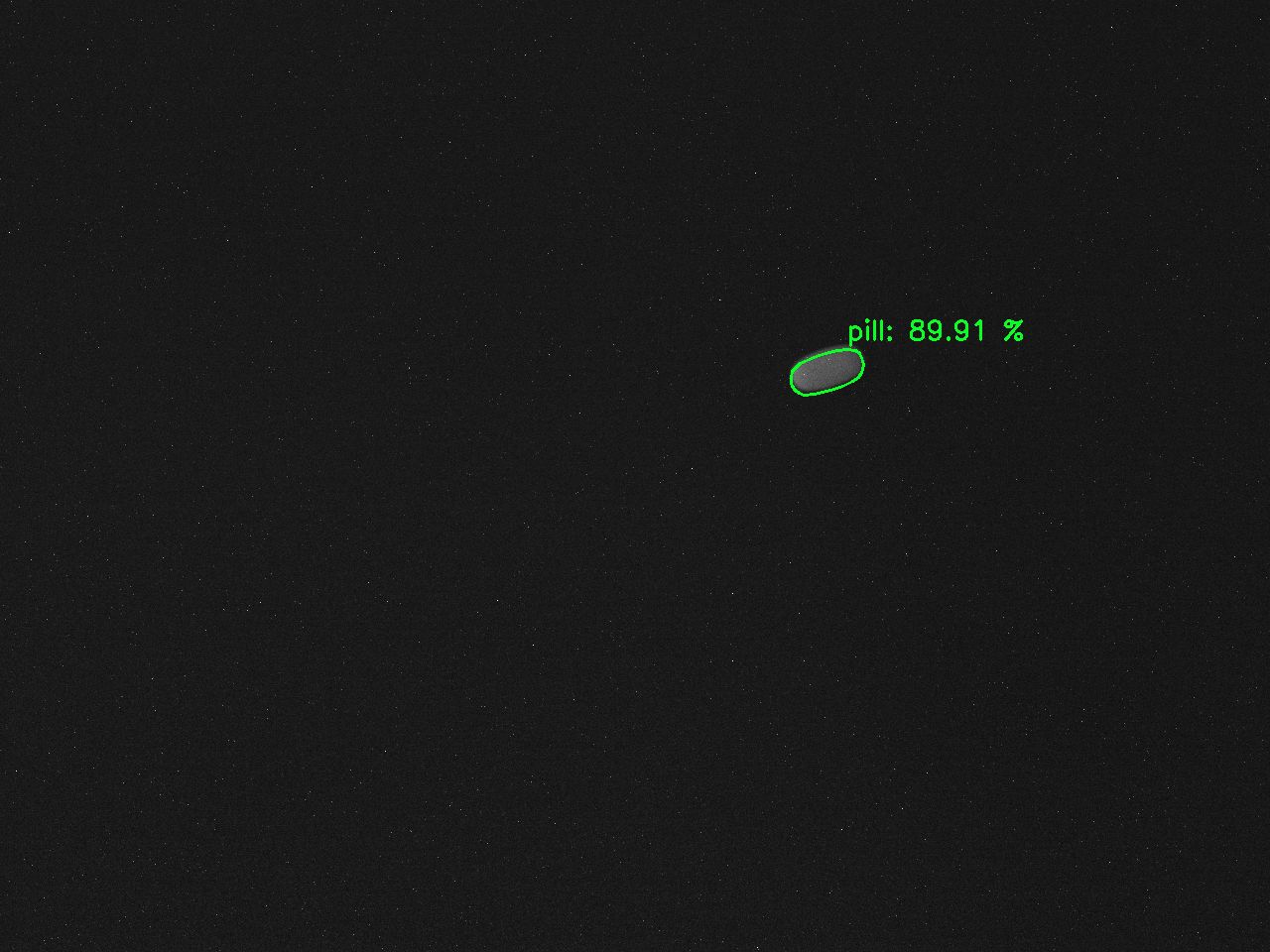}
         \caption{Band 4 output.}
         \label{fig:mask_out4}
     \end{subfigure}
     
     \hspace{1cm}
     
     \begin{subfigure}[b]{0.24\textwidth}
         \centering
         \includegraphics[trim={26cm 15cm 8cm 12cm},clip, width=\textwidth]{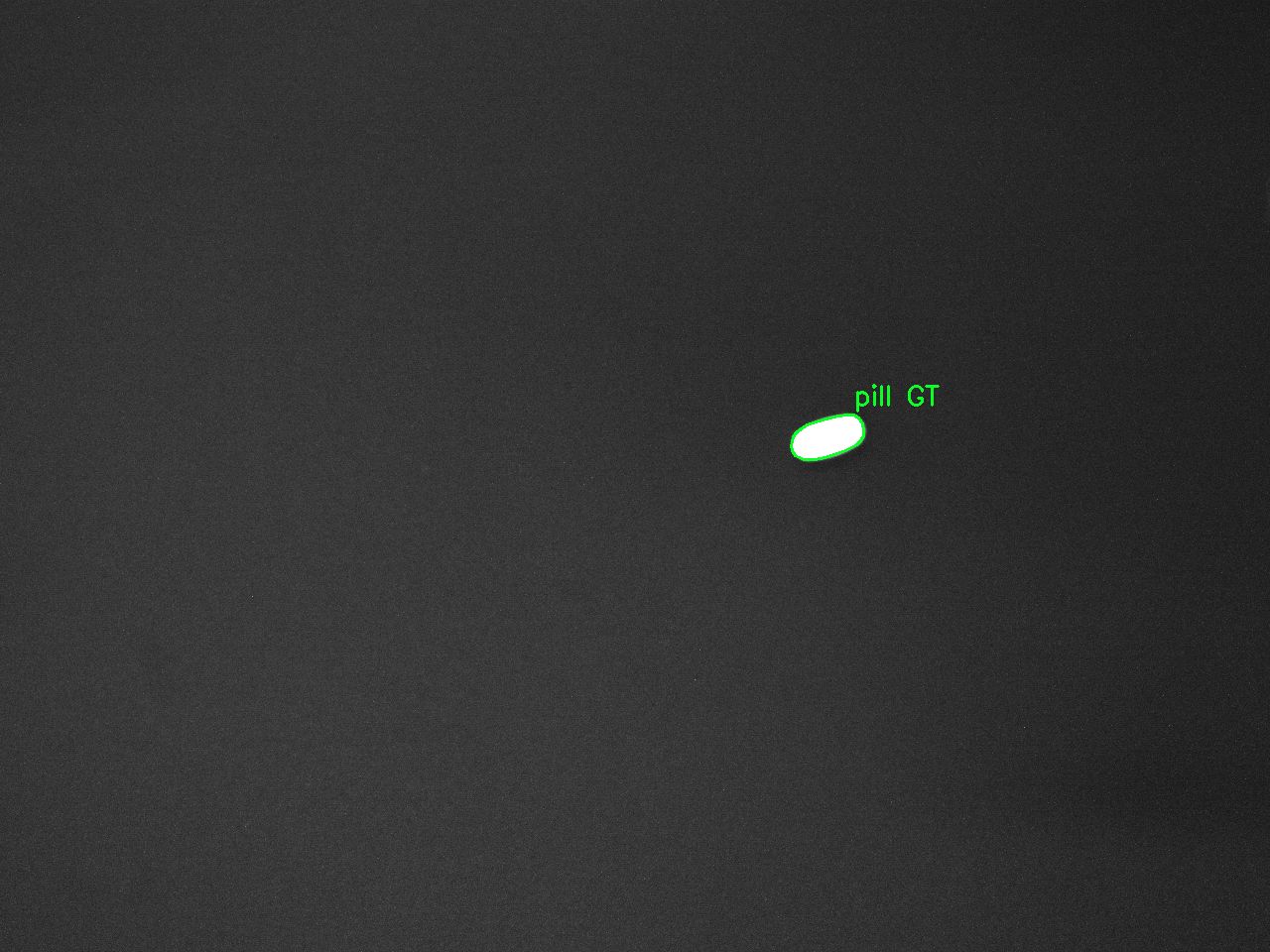}
         \caption{Band 1 reference.}
         \label{fig:mask_GT1}
     \end{subfigure}
     \begin{subfigure}[b]{0.24\textwidth}
         \centering
         \includegraphics[trim={30cm 15cm 4cm 12cm}, clip, width=\textwidth]{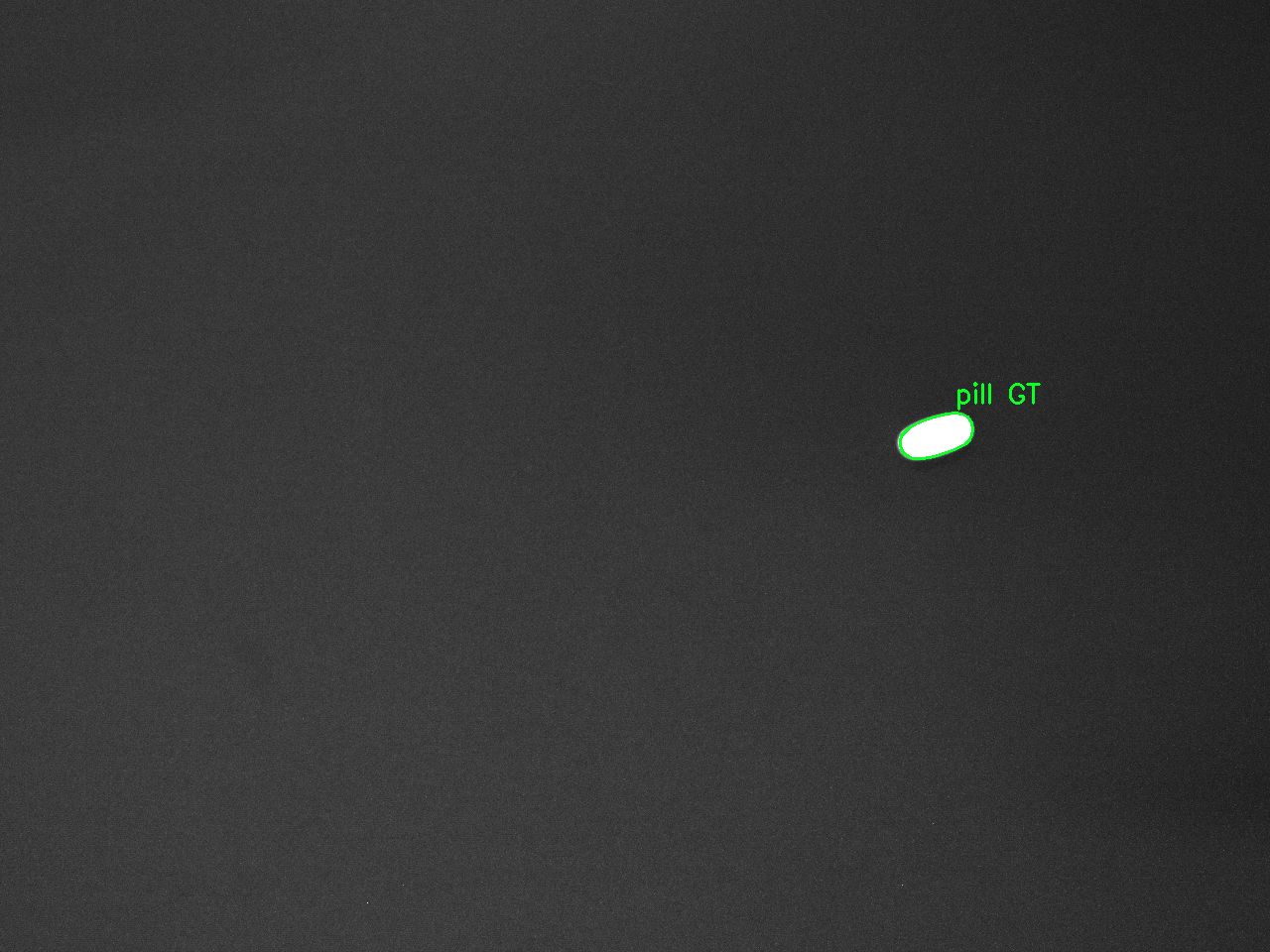}
         \caption{Band 2 reference.}
         \label{fig:mask_GT2}
     \end{subfigure}
     \begin{subfigure}[b]{0.24\textwidth}
         \centering
         \includegraphics[trim={29.5cm 17.5cm 4.5cm 9.5cm}, clip, width=\textwidth]{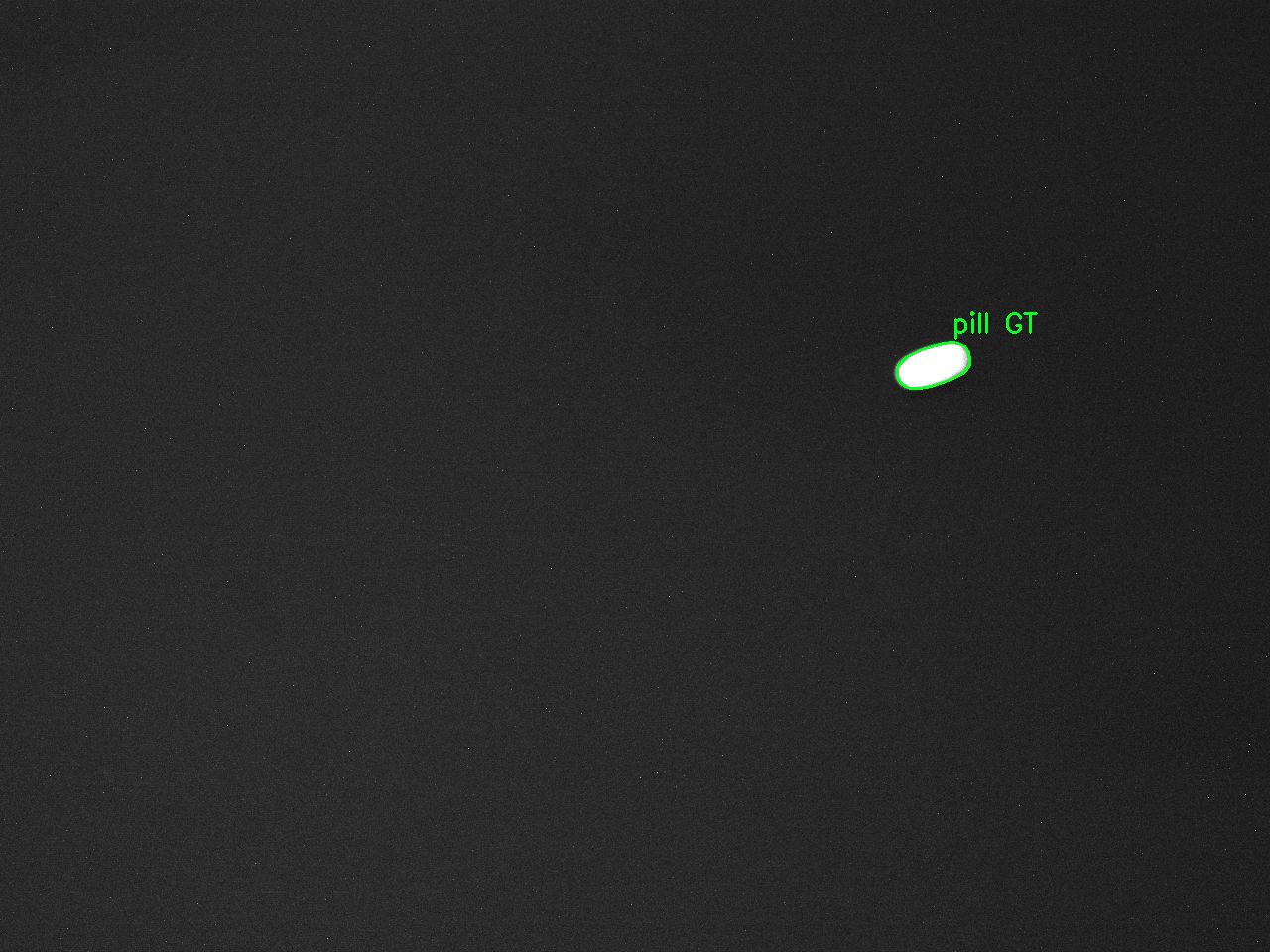}
         \caption{Band 3 reference.}
         \label{fig:mask_GT3}
     \end{subfigure}
     \begin{subfigure}[b]{0.24\textwidth}
         \centering
         \includegraphics[trim={25.5cm 17.5cm 8.5cm 9.5cm}, clip, width=\textwidth]{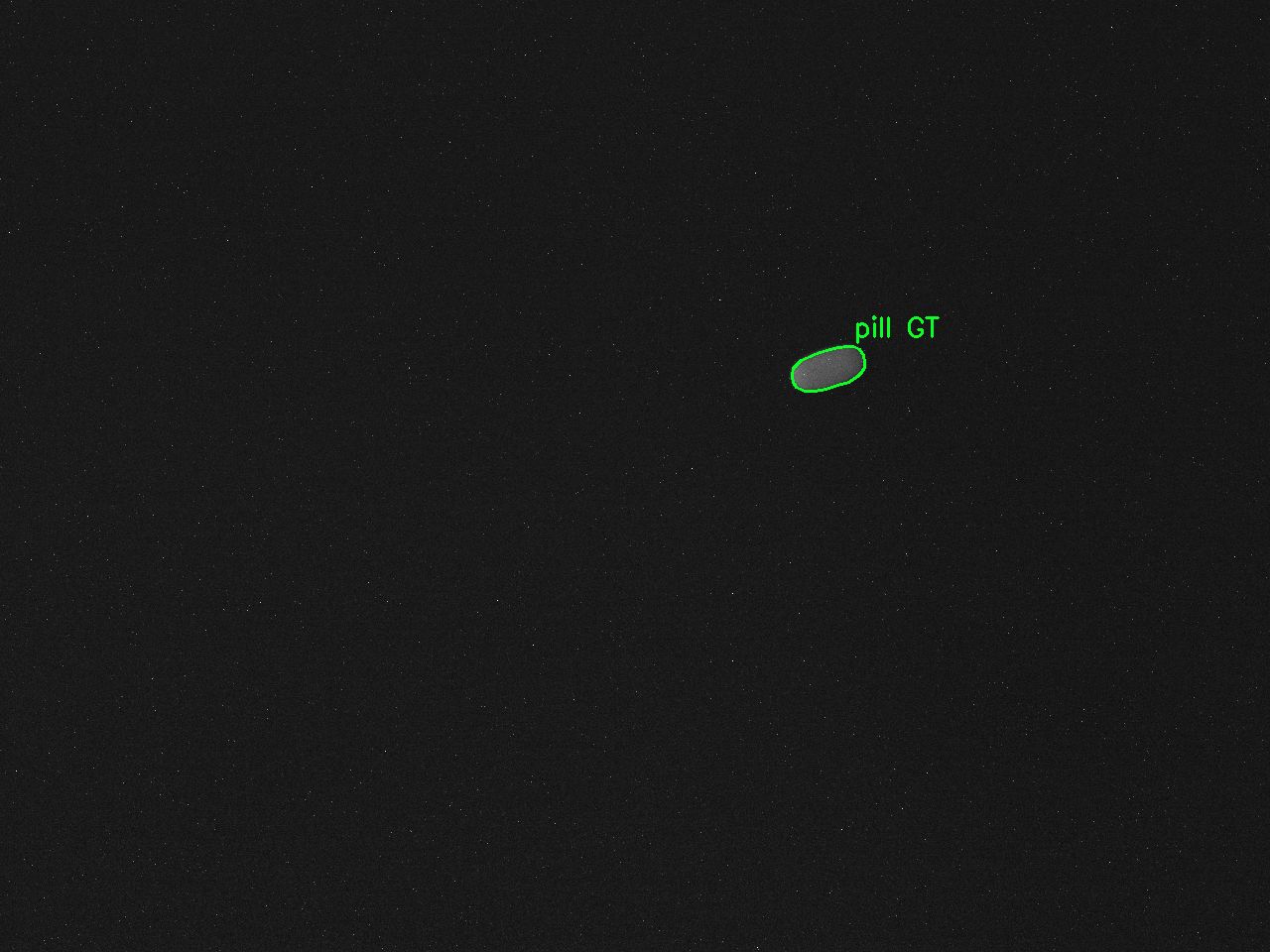}
         \caption{Band 4 reference.}
         \label{fig:mask_GT4}
     \end{subfigure}

     \hspace{1cm}
     
     \caption{Mask labeled experiment: (a) reference image (band 5) to start with, (b, c, d, e) transformed labels (bands 1-4) and (f, g, h, i) ground truth labels for comparison purposes (bands 1-4).}\label{fig:transfResMask}
\end{figure*}

Moving onto a more difficult problem, mask labeled images are tested. These images are transformed using matrices $^1_5T_{MK}$, $^2_5T_{MK}$, $^3_5T_{MK}$ and $^4_5T_{MK}$ from Table \ref{tab:maskTransform}. Results of the process are shown in Fig. \ref{fig:transfResMask}. The layout follows the same pattern as in Fig. \ref{fig:transfResBB}. Taking that into account, the worst result is again band 4, being the reason the same as in the previous case. The pill starts to disappear (compared to the other 3 bands). However, the labeling process is still successful. 

\subsection{RGB label transferability}
\label{subsec:res_rgbTransfer}
Once the labeling process is shown to be successful, a new experiment will be performed - labeling in RGB images created from the multispectral images and then move these labels into the other bands. 

For that purpose, bands representing the red, green and blue frequencies need to be mixed together. According to Table \ref{tab:wavelengths}, bands 1-3 from the red module are the ones required. So the images are converted from bands 1-3 to band 5 by using Eq. \ref{eq:bands2RGB}, obtaining an artificial RGB image $im_{RGB} = [R,G,B]$. 

\begin{equation}
\label{eq:bands2RGB}
\begin{split}
    R = _{5}^{3}\textrm{T}_{BB | mask}^{-1} \cdot im_{band 3} \\ 
    G = _{5}^{2}\textrm{T}_{BB | mask}^{-1} \cdot im_{band 2} \\
    B = _{5}^{1}\textrm{T}_{BB | mask}^{-1} \cdot im_{band 1}
\end{split}
\end{equation}

Then, this fake RGB image is labeled in \gls{bb} or mask format $l_{RGB} = [(l_{RGB: 1x},l_{RGB: 1y}), ... , (l_{RGB: nx},l_{RGB: ny}), ..., \\ (l_{RGB: Nx},l_{RGB: Ny})]$ by the user. Once the label is ready, it is transferred back into the other bands in the camera, obtaining labels in all frequencies (Eq. \ref{eq:RGB2bands}).

\begin{equation}
\label{eq:RGB2bands}
\begin{split}
    l_{band 1} = _{5}^{1}\textrm{T}_{BB | mask} \cdot l_{RGB} \\  
    l_{band 2} = _{5}^{2}\textrm{T}_{BB | mask} \cdot l_{RGB} \\
    l_{band 3} = _{5}^{3}\textrm{T}_{BB | mask} \cdot l_{RGB} \\
    l_{band 4} = _{5}^{4}\textrm{T}_{BB | mask} \cdot l_{RGB} \\
    l_{band 5} = I_{2x3} \cdot l_{RGB}
\end{split}
\end{equation}

\begin{figure*}[!h]
\centering
    \begin{subfigure}[b]{0.32\textwidth}
         \centering
         \includegraphics[width=\textwidth]{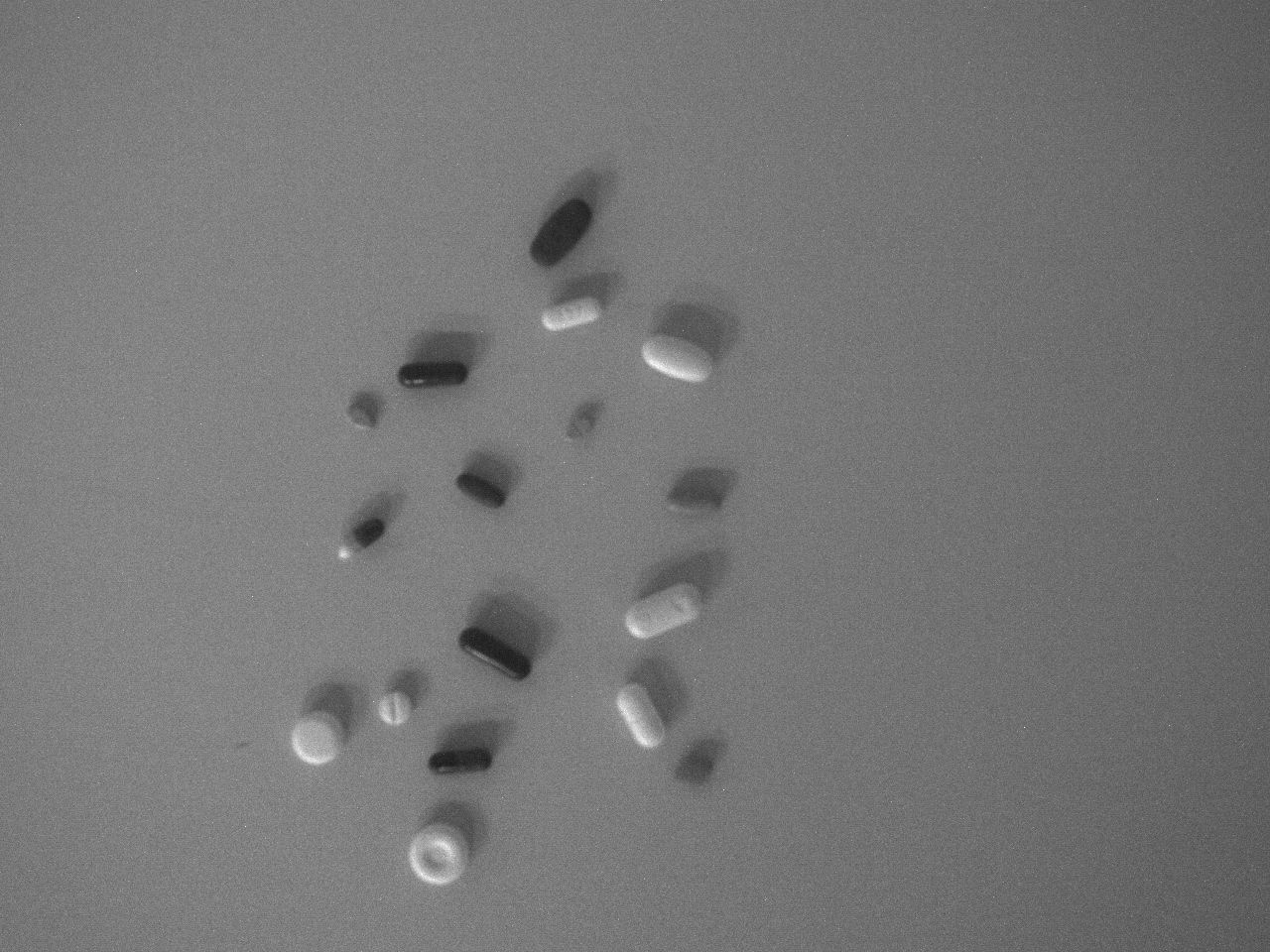}
         \caption{Band 1: blue.}
         \label{fig:exp2RGBCreate_blue}
    \end{subfigure}
    \begin{subfigure}[b]{0.32\textwidth}
         \centering
         \includegraphics[width=\textwidth]{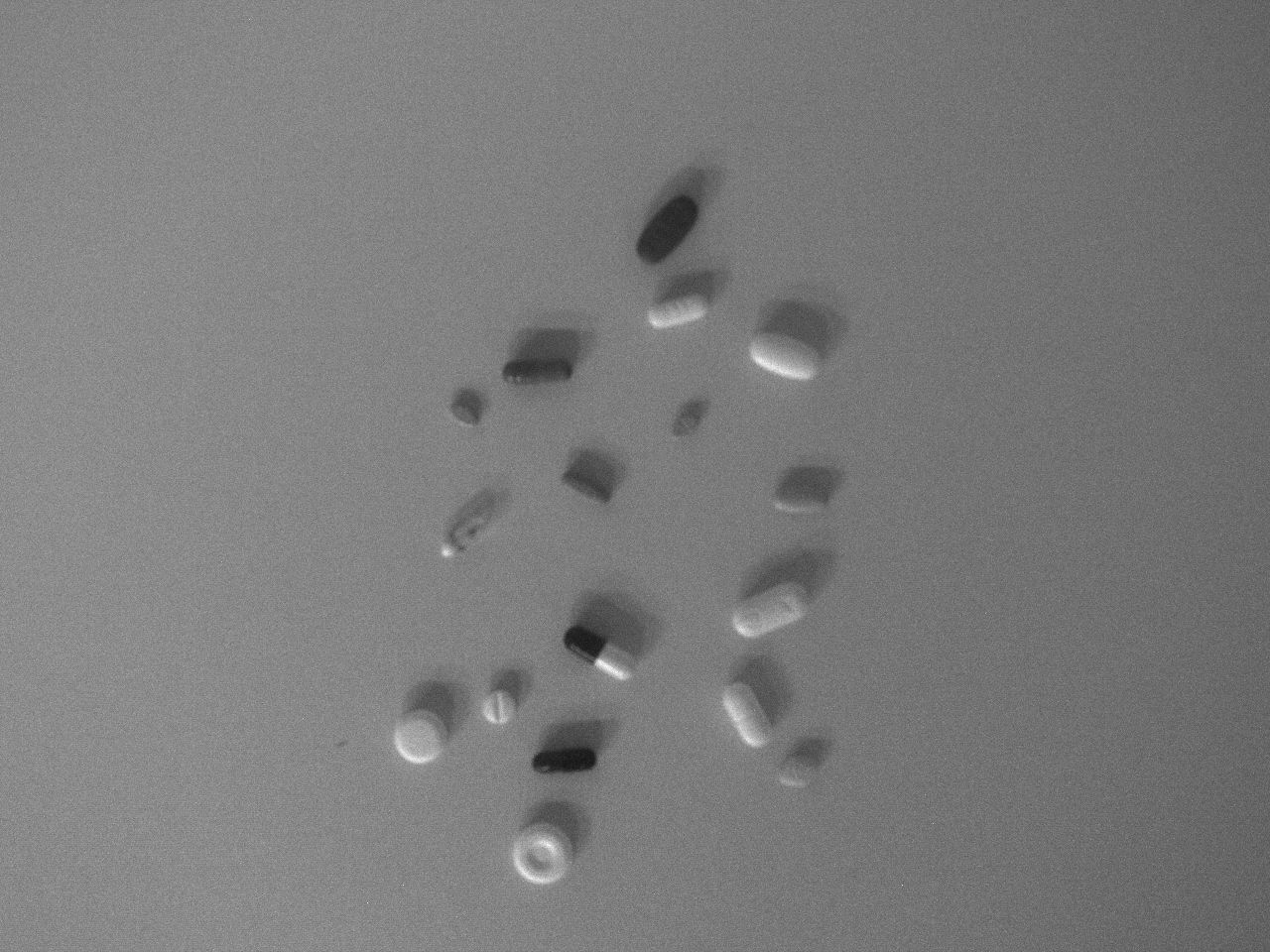}
         \caption{Band 2: green.}
         \label{fig:exp2RGBCreate_green}
    \end{subfigure}
    \begin{subfigure}[b]{0.32\textwidth}
         \centering
         \includegraphics[width=\textwidth]{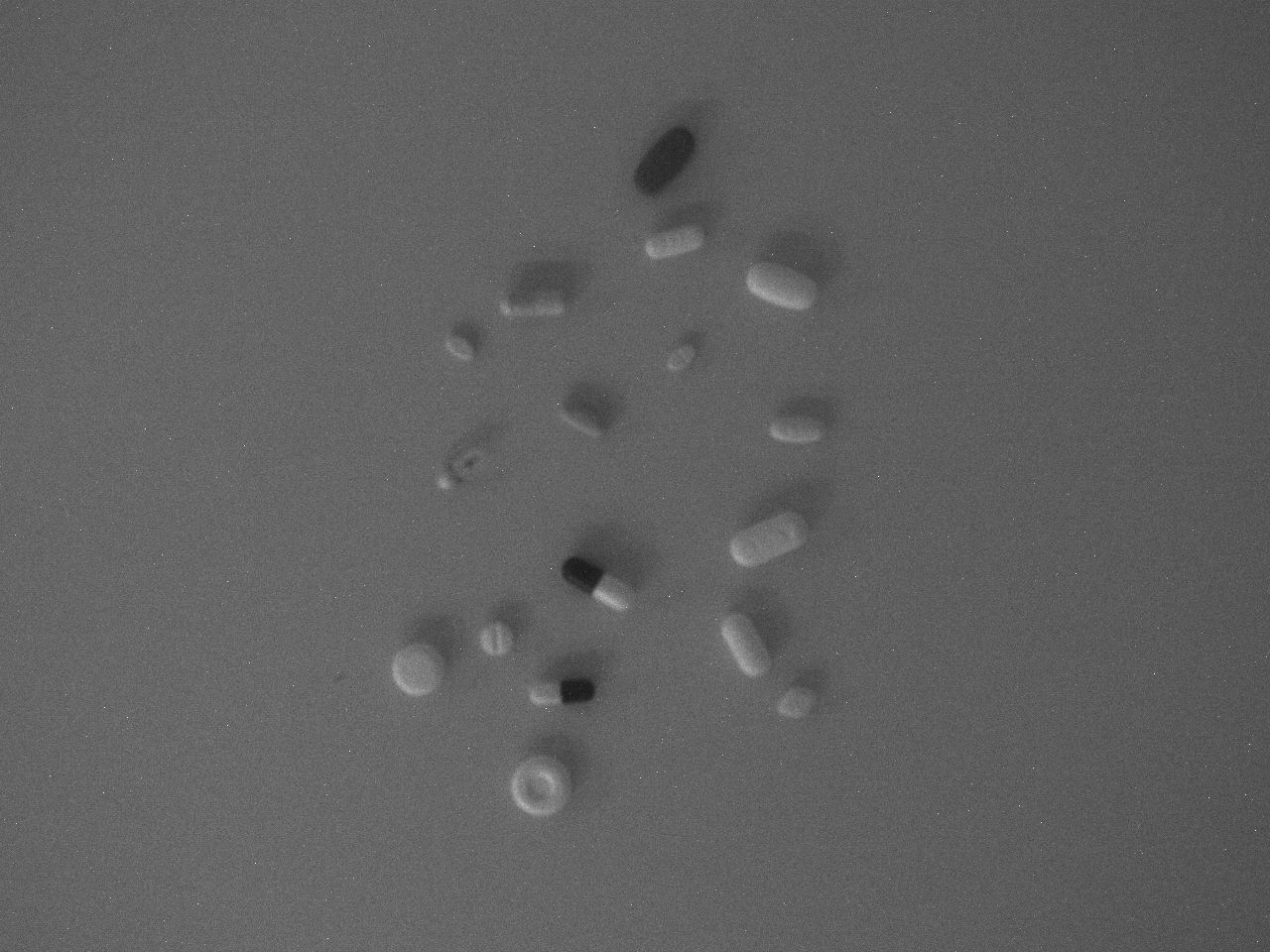}
         \caption{Band 3: red.}
         \label{fig:exp2RGBCreate_red}
     \end{subfigure}
     
     \hspace{1cm}
     
     \centering
     \begin{subfigure}[b]{0.32\textwidth}
         \centering
         \includegraphics[width=\textwidth]{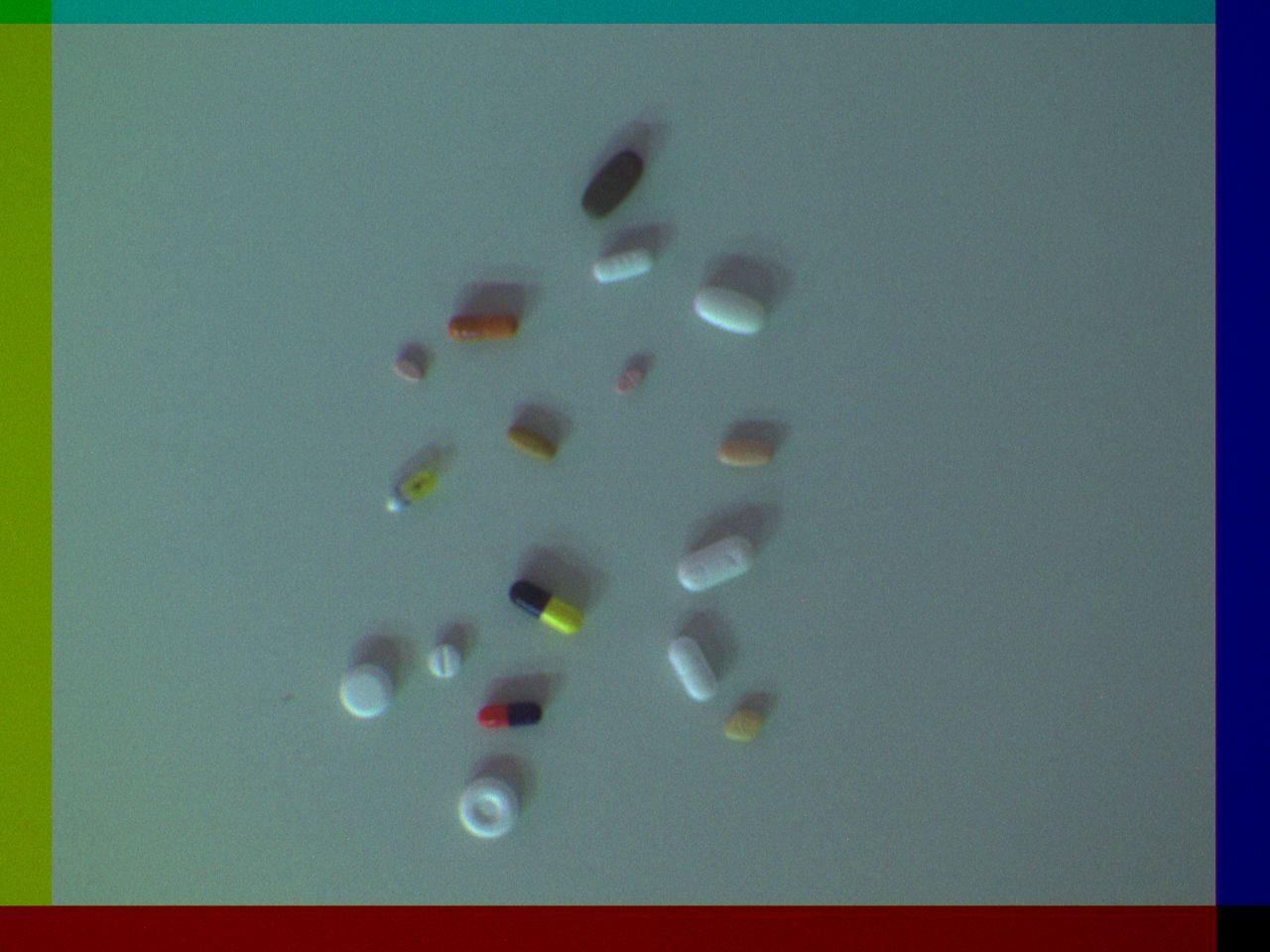}
         \caption{Fake RGB image.}
         \label{fig:exp2RGBCreate_RGB}
     \end{subfigure}

     \hspace{1cm}
     
     \caption{Combination of 3 channels to create fake RGB image: (a,b,c) bands 1, 2, 3 respectively, (d) Generated fake RGB image.}\label{fig:exp2RGBCreate}
\end{figure*}

The first process for generating the $im_{RGB} = [R,G,B]$ can be seen in Fig. \ref{fig:exp2RGBCreate}. For it, a combination of blue (Fig. \ref{fig:exp2RGBCreate_blue}), green (Fig. \ref{fig:exp2RGBCreate_green}) and red (Fig. \ref{fig:exp2RGBCreate_red}) images is performed, generating a fake RGB image (Fig. \ref{fig:exp2RGBCreate_RGB}).

\begin{figure*}[!h]
\centering
    \begin{subfigure}[b]{0.32\textwidth}
         \centering
         \includegraphics[trim={10cm 2.5cm 15cm 2.5cm}, clip, width=\textwidth]{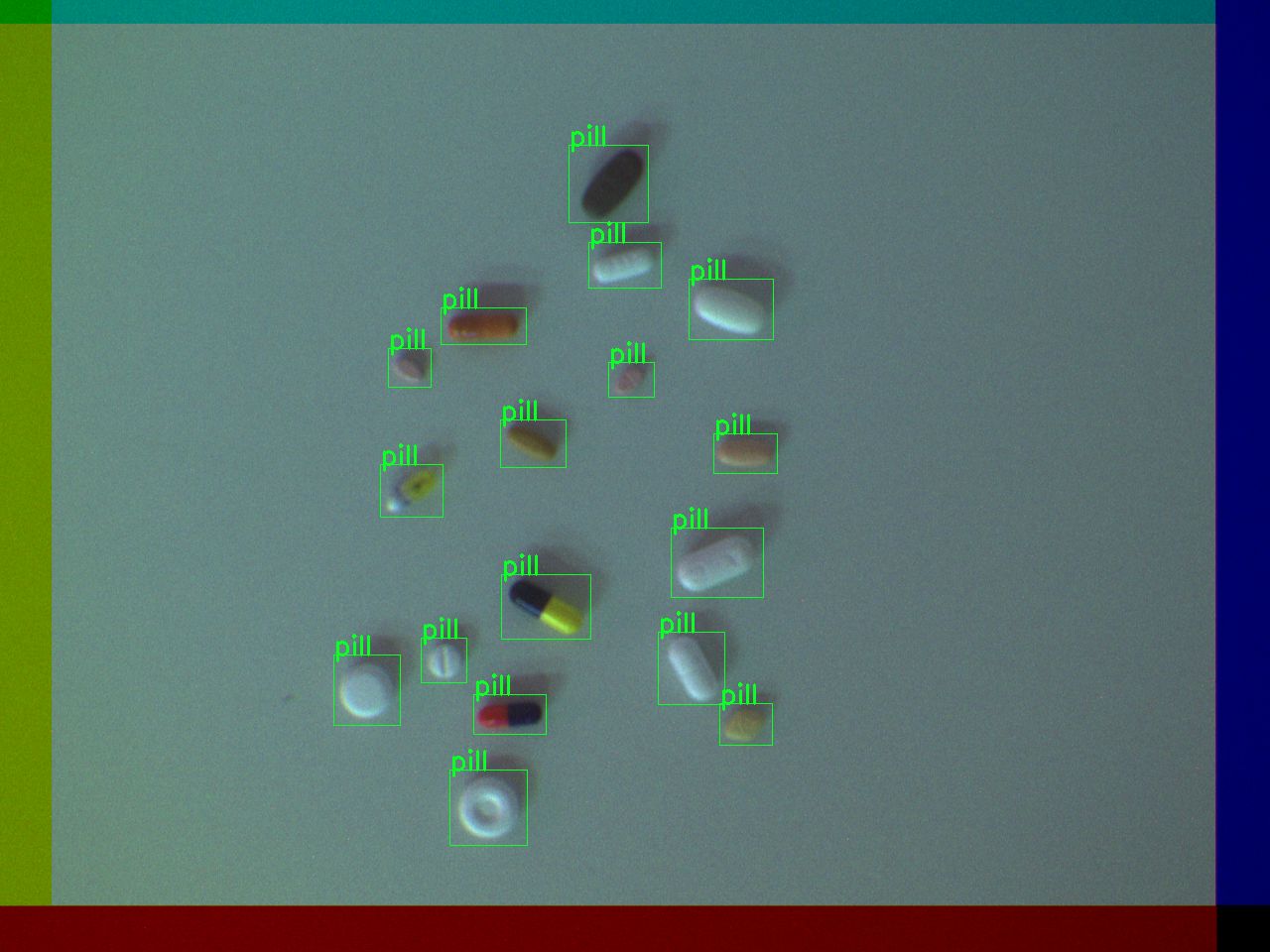}
         \caption{Fake RGB labeled with BBs.}
         \label{fig:exp2BBResults_RGB}
     \end{subfigure}     
     \begin{subfigure}[b]{0.32\textwidth}
         \centering
         \includegraphics[trim={7.5cm 1cm 17.5cm 4cm}, clip, width=\textwidth]{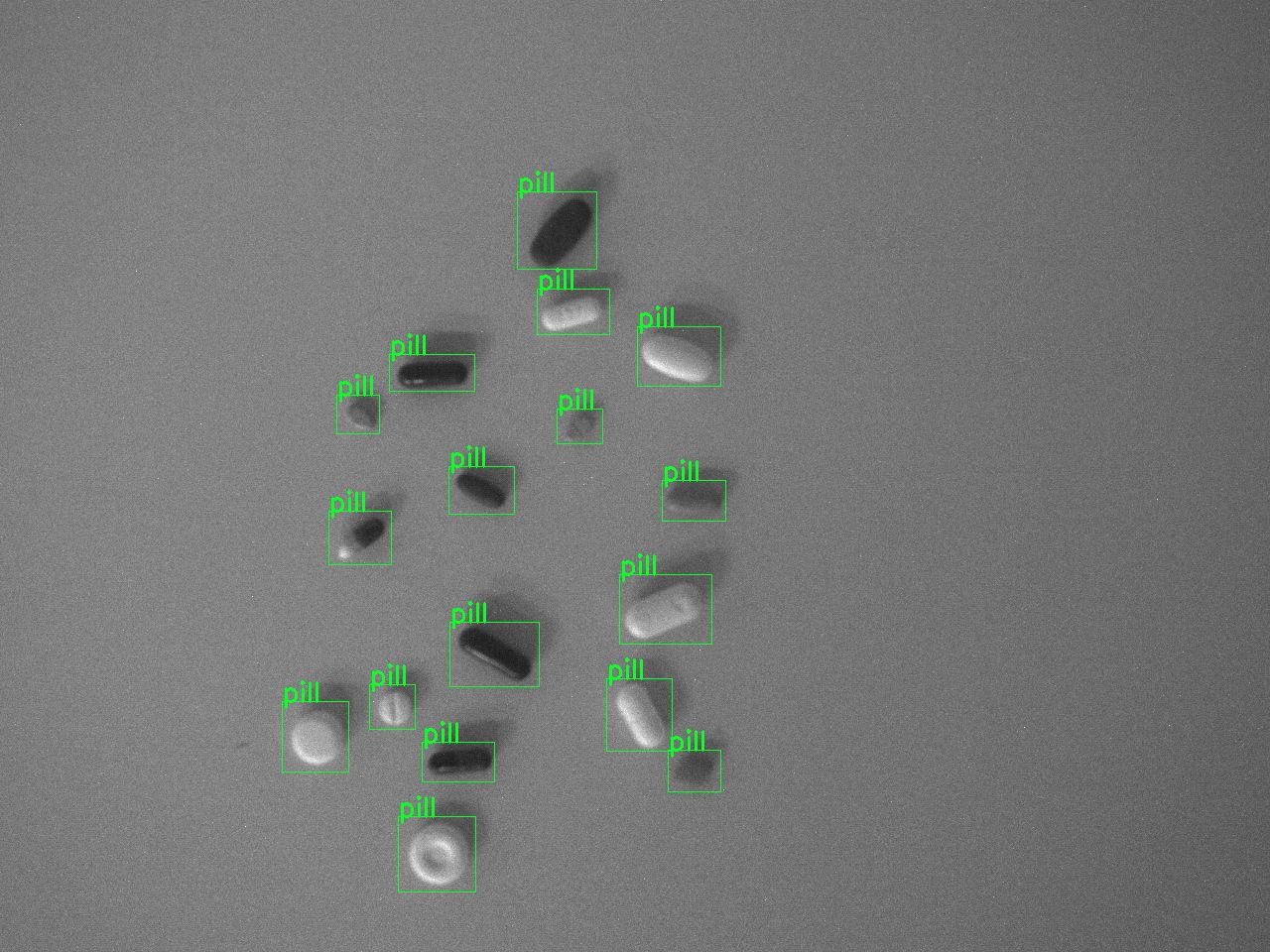}
         \caption{Band 1 with transfered labels.}
         \label{fig:exp2BBResults_b1}
     \end{subfigure}
     \begin{subfigure}[b]{0.32\textwidth}
         \centering
         \includegraphics[trim={11.5cm 1.5cm 13.5cm 3.5cm}, clip, width=\textwidth]{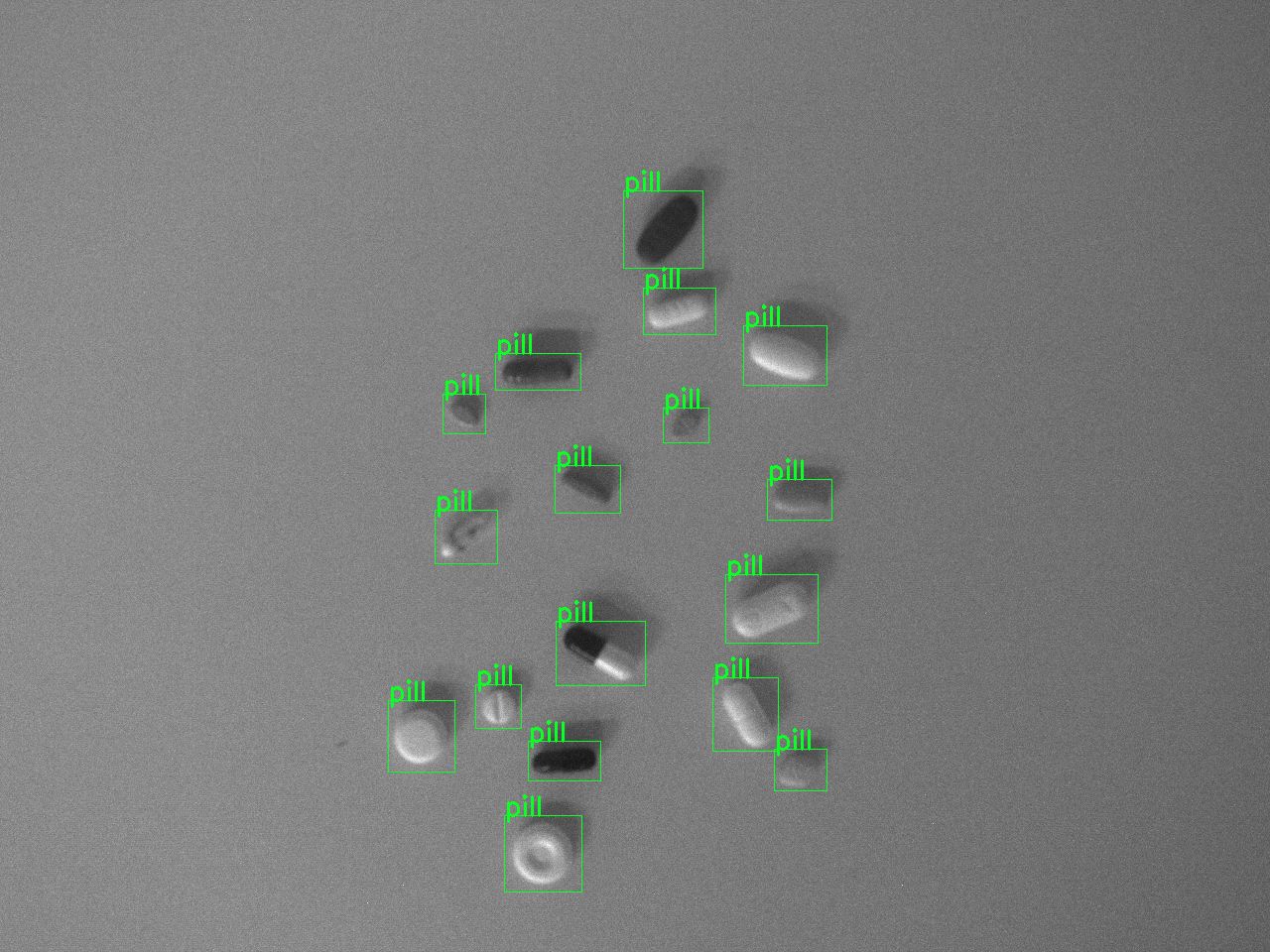}
         \caption{Band 2 with transfered labels.}
         \label{fig:exp2BBResults_b2}
     \end{subfigure}
     
     \hspace{1cm}
     
     \begin{subfigure}[b]{0.32\textwidth}
         \centering
         \includegraphics[trim={11.5cm 3cm 13.5cm 2cm}, clip, width=\textwidth]{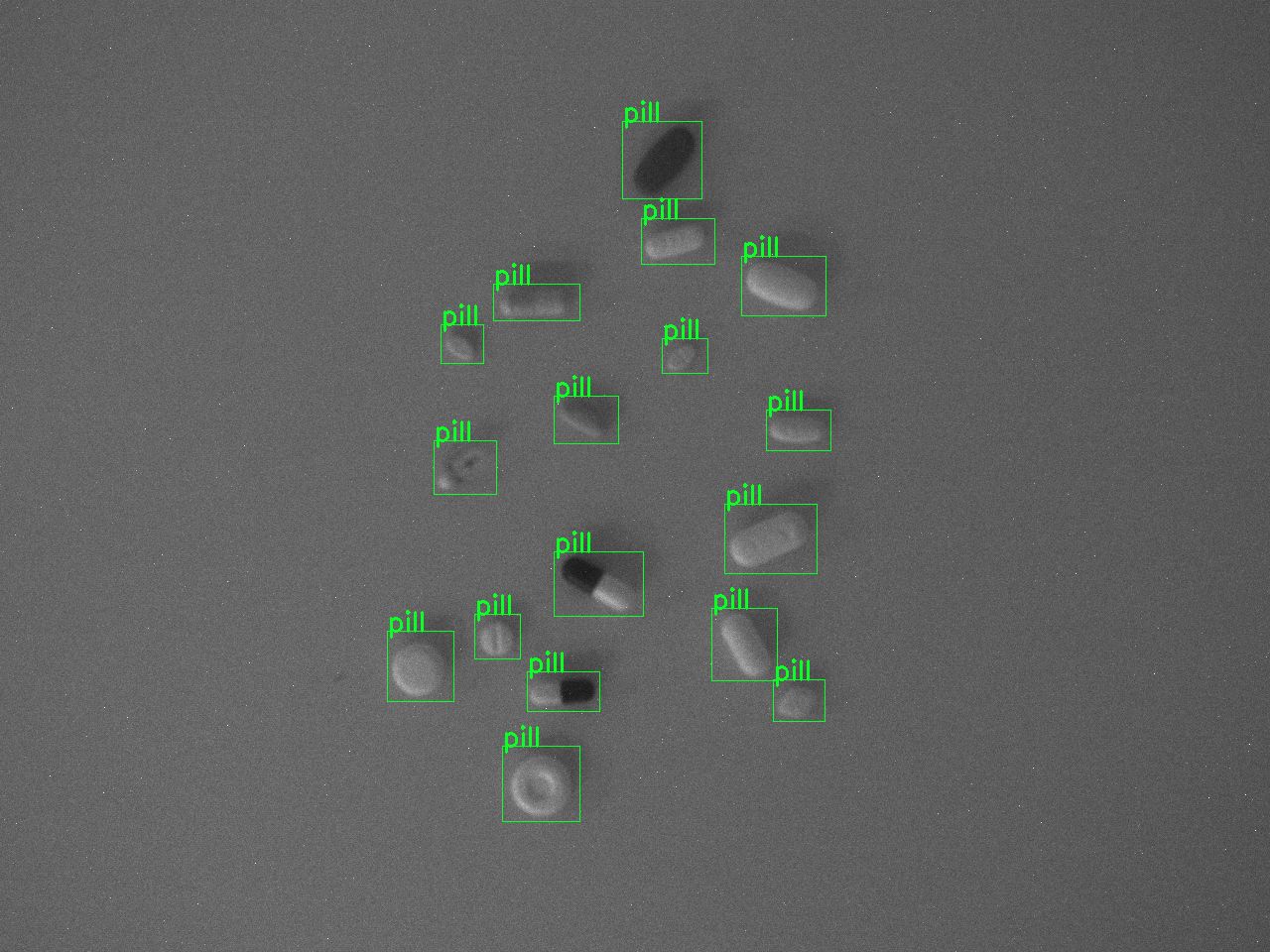}
         \caption{Band 3 with transfered labels.}
         \label{fig:exp2BBResults_b3}
     \end{subfigure}     
     \begin{subfigure}[b]{0.32\textwidth}
         \centering
         \includegraphics[trim={8cm 3cm 17cm 2cm}, clip, width=\textwidth]{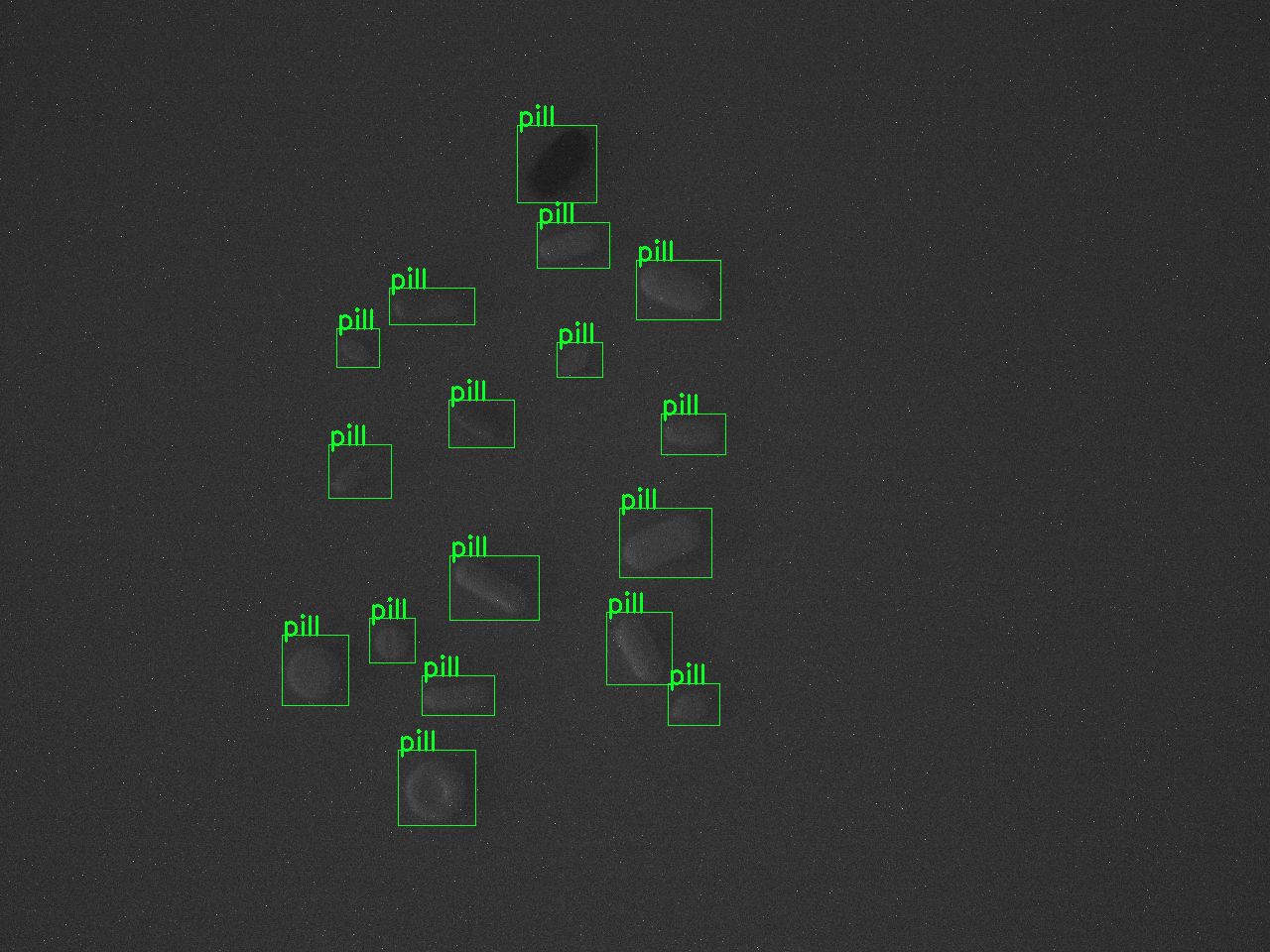}
         \caption{Band 4 with transfered labels.}
         \label{fig:exp2BBResults_b4}
     \end{subfigure}
    \begin{subfigure}[b]{0.32\textwidth}
         \centering
         \includegraphics[trim={9cm 2.5cm 16cm 2.5cm}, clip, width=\textwidth]{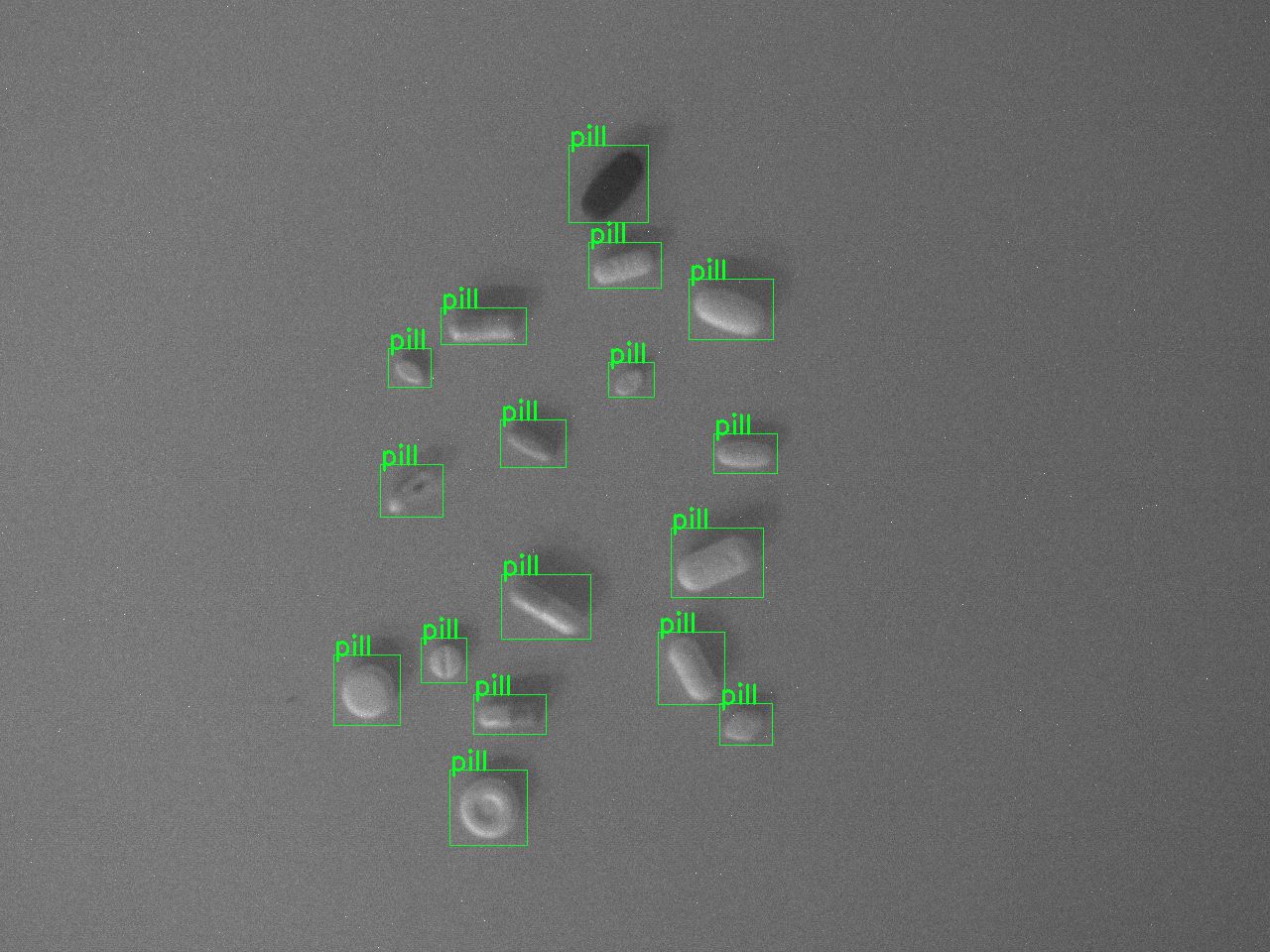}
         \caption{Band 5 with transfered labels.}
         \label{fig:exp2BBResults_b5}
     \end{subfigure}
     
     \hspace{1cm}
          
     \caption{BB labeled fake RGB image and transfered labels: (a) Fake RGB image with \gls{bb} labels, (b) labels transfered to band 1, (c) labels transfered to band 2, (d) labels transfered to band 3, (e) labels transfered to band 4 and (f) labels transfered to band 5.}\label{fig:exp2BBResults}
\end{figure*}

\begin{figure*}[!h]
\centering
    \begin{subfigure}[b]{0.32\textwidth}
         \centering
         \includegraphics[trim={10cm 2.5cm 15cm 2.5cm}, clip, width=\textwidth]{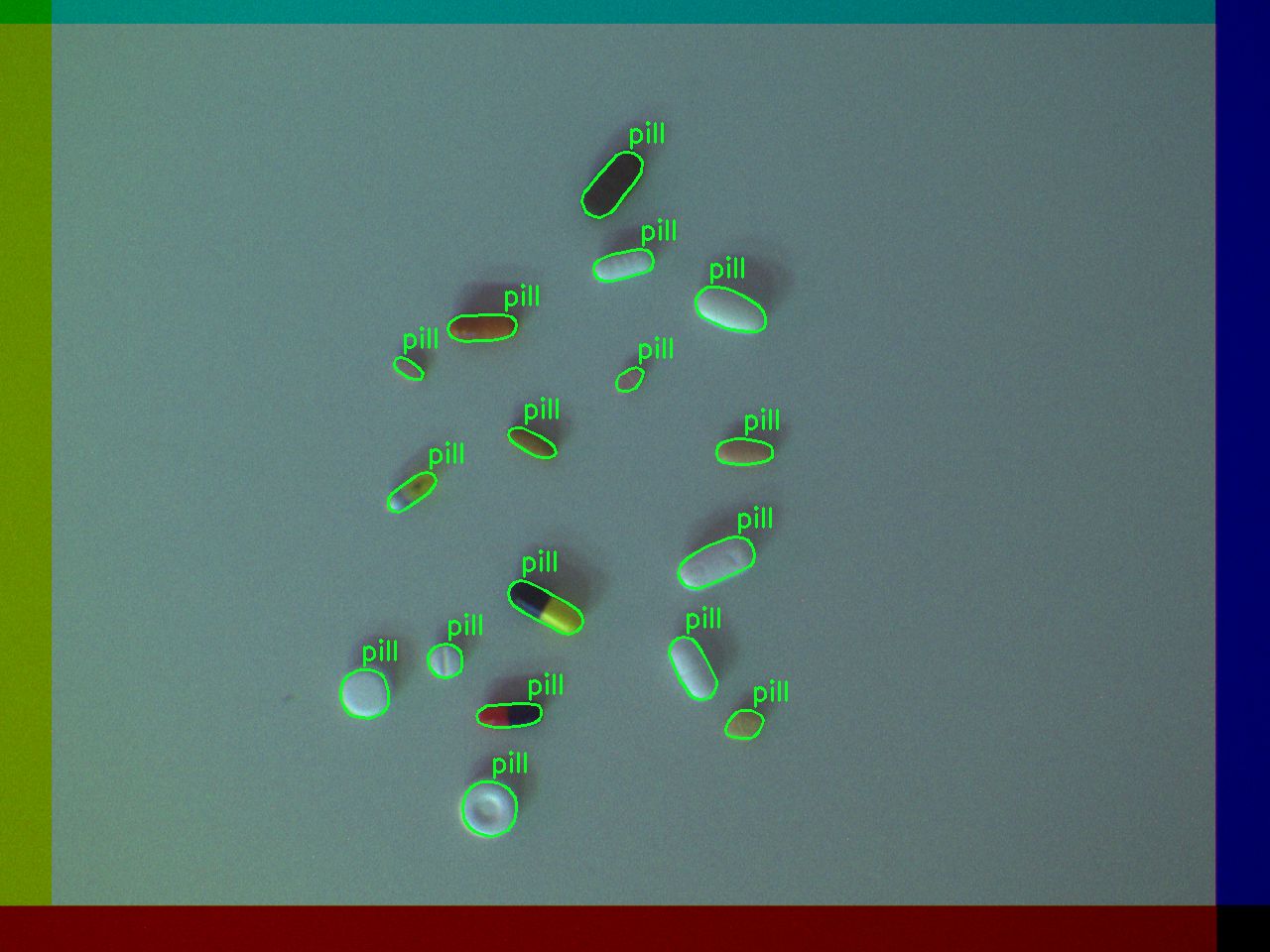}
         \caption{Fake RGB labeled with masks.}
         \label{fig:exp2MaskResults_RGB}
     \end{subfigure}     
     \begin{subfigure}[b]{0.32\textwidth}
         \centering
         \includegraphics[trim={7.5cm 1cm 17.5cm 4cm}, clip, width=\textwidth]{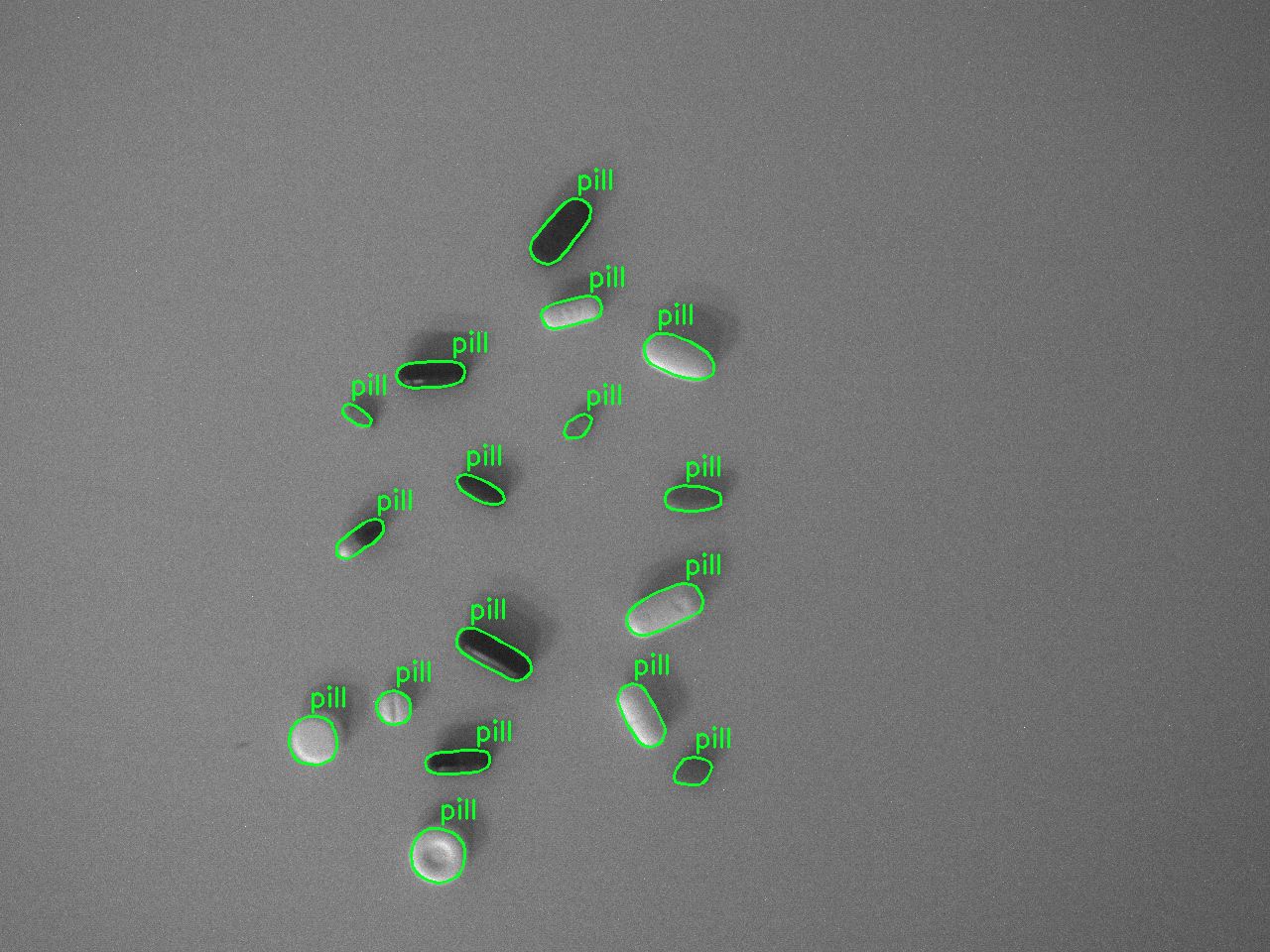}
         \caption{Band 1 with transfered labels.}
         \label{fig:fig:exp2MaskResults_b1}
     \end{subfigure}
     \begin{subfigure}[b]{0.32\textwidth}
         \centering
         \includegraphics[trim={11.5cm 1.5cm 13.5cm 3.5cm}, clip, width=\textwidth]{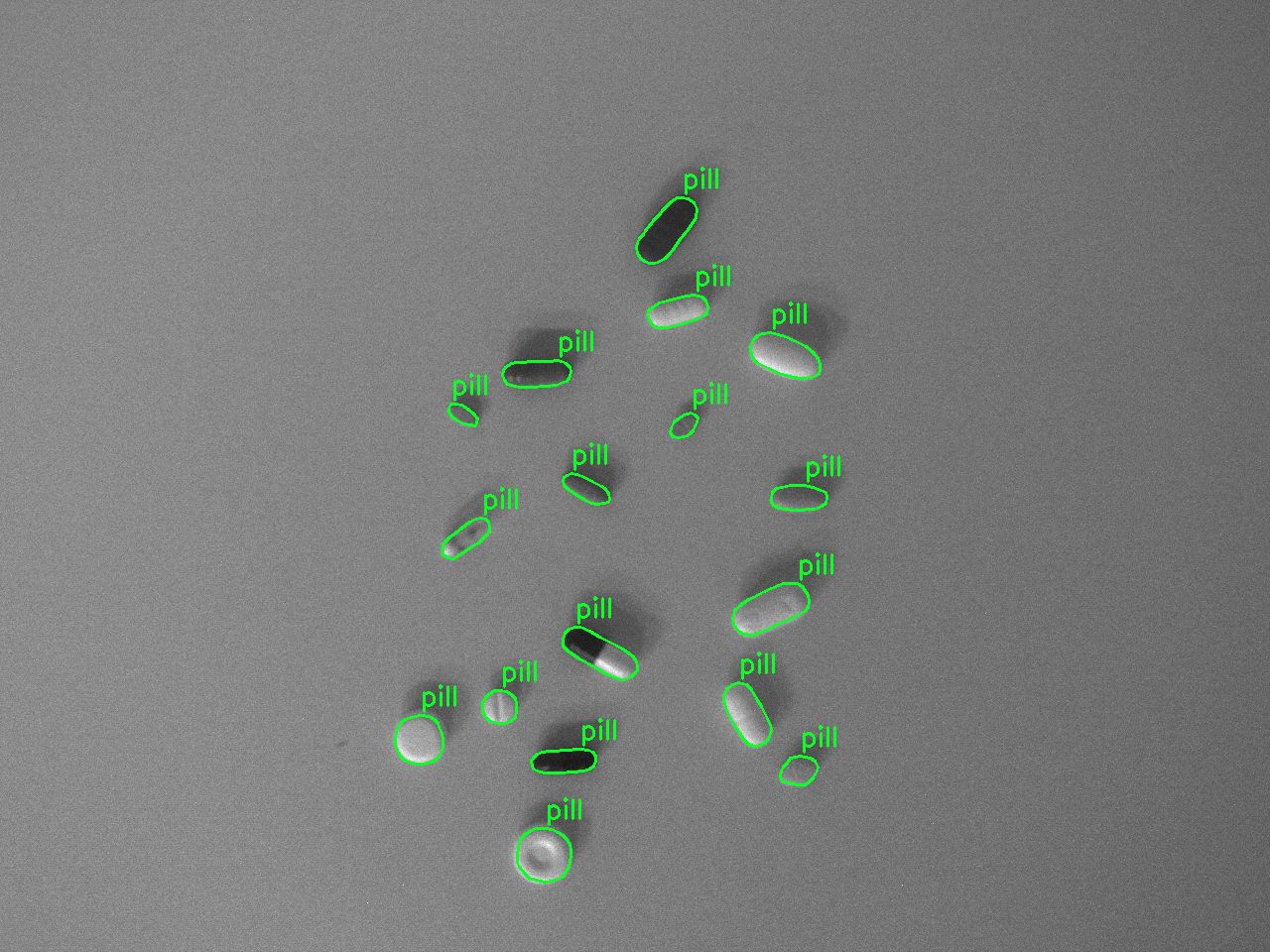}
         \caption{Band 2 with transfered labels.}
         \label{fig:exp2MaskResults_b2}
     \end{subfigure}
     
     \hspace{1cm}
     
     \begin{subfigure}[b]{0.32\textwidth}
         \centering
         \includegraphics[trim={11.5cm 3cm 13.5cm 2cm}, clip, width=\textwidth]{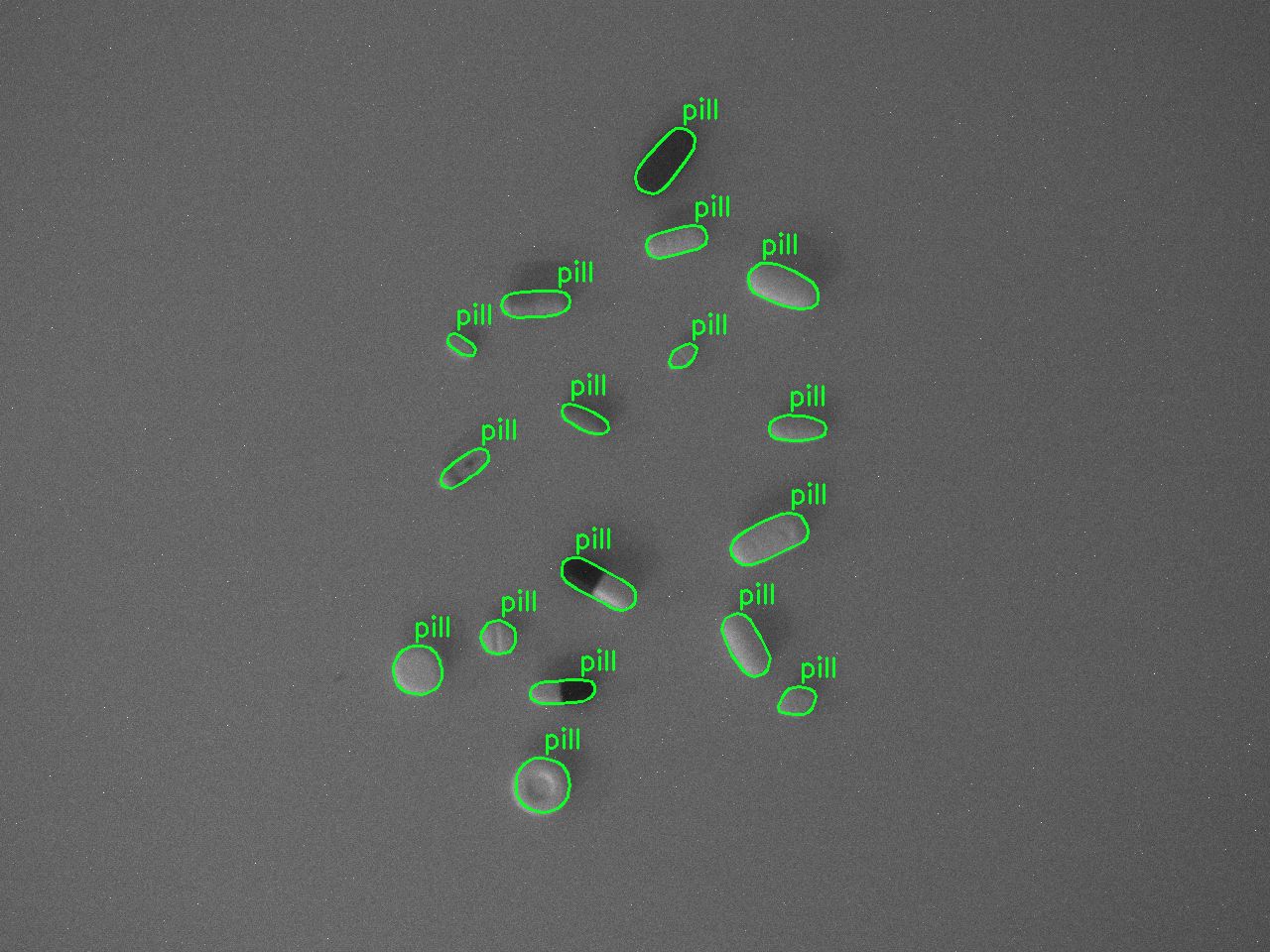}
         \caption{Band 3 with transfered labels.}
         \label{fig:exp2MaskResults_b3}
     \end{subfigure}     
     \begin{subfigure}[b]{0.32\textwidth}
         \centering
         \includegraphics[trim={8cm 3cm 17cm 2cm}, clip, width=\textwidth]{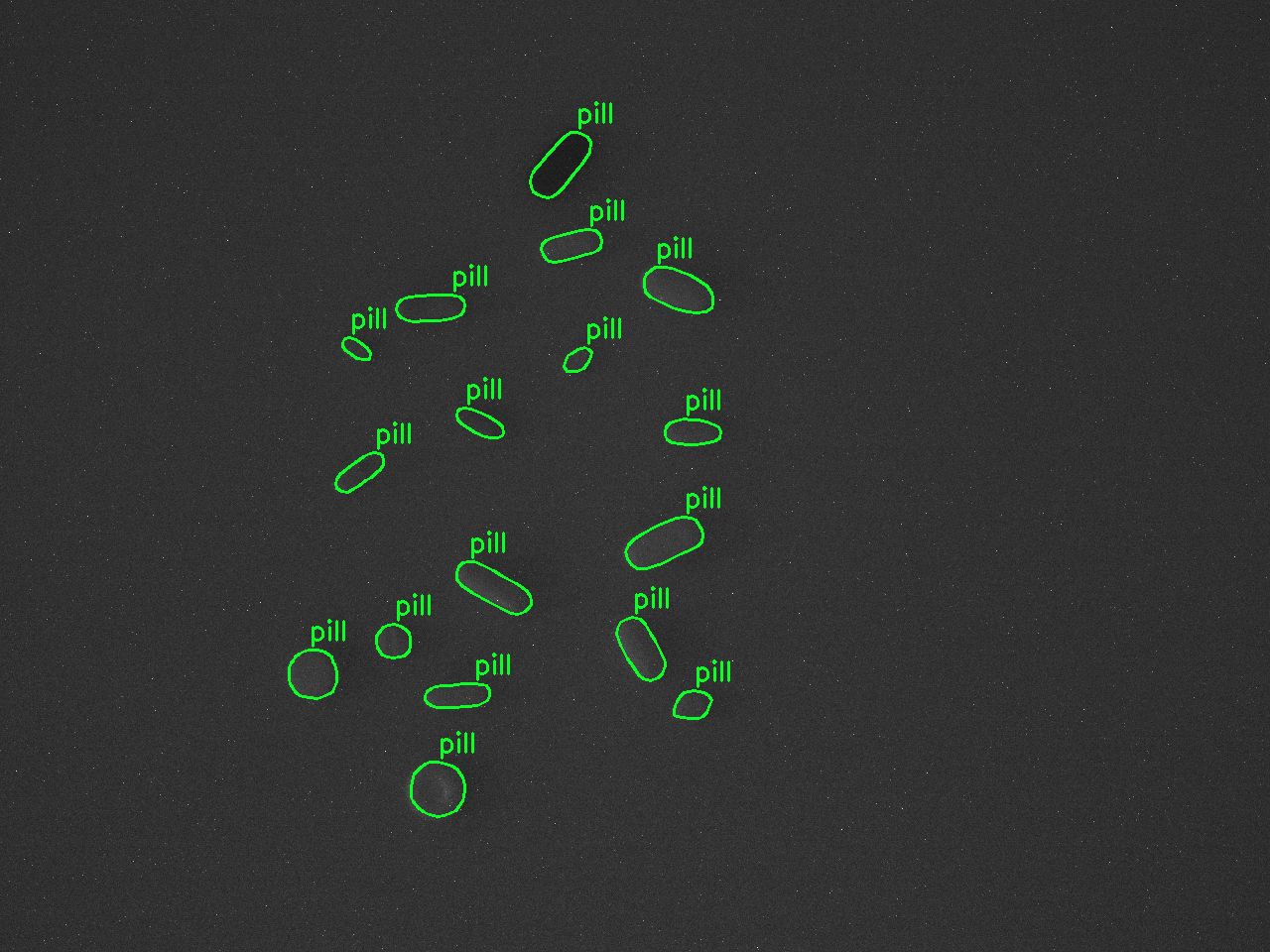}
         \caption{Band 4 with transfered labels.}
         \label{fig:exp2MaskResults_b4}
     \end{subfigure}
    \begin{subfigure}[b]{0.32\textwidth}
         \centering
         \includegraphics[trim={9cm 2.5cm 16cm 2.5cm}, clip, width=\textwidth]{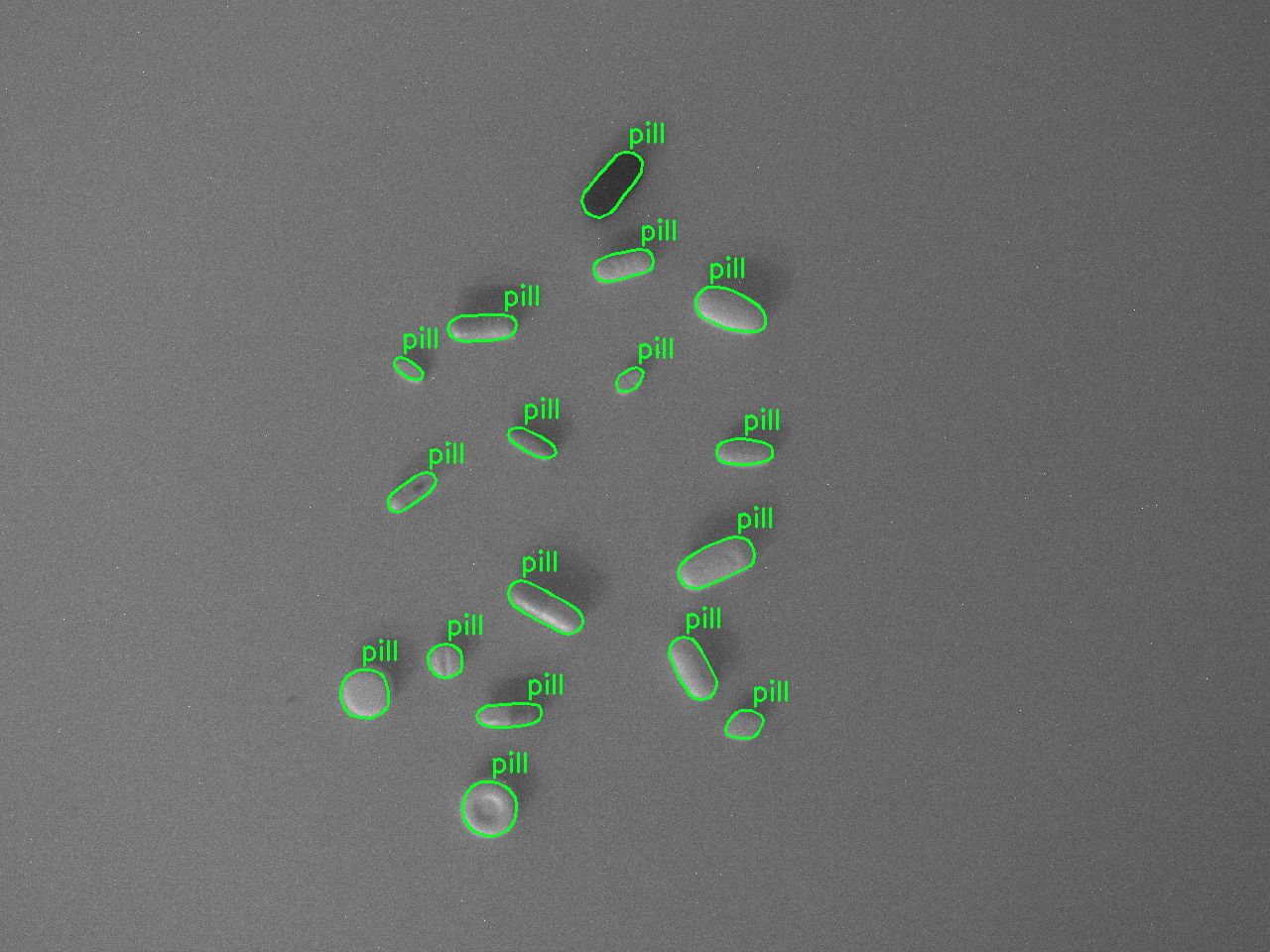}
         \caption{Band 5 with transfered labels.}
         \label{fig:exp2MaskResults_b5}
     \end{subfigure}
     
     \hspace{1cm}
    
     \caption{Mask labeled fake RGB image and transfered labels: (a) Fake RGB image with mask labels, (b) labels transfered to band 1, (c) labels transfered to band 2, (d) labels transfered to band 3, (e) labels transfered to band 4 and (f) labels transfered to band 5.}\label{fig:exp2MaskResults}         
\end{figure*}

Once the fake RGB image is generated, objects have been labeled in \gls{bb} (Fig. \ref{fig:exp2BBResults_RGB}) and mask (Fig. \ref{fig:exp2MaskResults_RGB}) formats. After applying the transformations obtained from Section \ref{subsec:res_transformations}, the images with \gls{bb} (Figs. \ref{fig:exp2BBResults_b1}-\ref{fig:exp2BBResults_b5}) and masks (Figs. \ref{fig:fig:exp2MaskResults_b1}-\ref{fig:exp2MaskResults_b5}) were obtained. As it can be seen, obtained results convincingly demonstrate the accuracy of the proposed approach. For example, objects in band 4 with \gls{bb} (Fig. \ref{fig:exp2BBResults_b4}) as well as with mask (Fig. \ref{fig:exp2MaskResults_b4}) labeling disappear partially. However, due to the designed method, objects in those positions are labeled even though they are not visible.

\section{\uppercase{Conclusion}}
\label{sec:conclusion}
In this work, a method for labeling automatically multispectral images starting with a single band image \gls{bb} or mask labeled is presented. 

In order to achieve it, a process with two steps is used: phase correlation and refinement. During the first one, the transformation between two images is obtained by smoothing the image with a Hanning window, transforming the spatial domain images into the frequency domain with the Fourier discrete transform, applying the cross-power spectrum formula to retain just the phase information of the images, converting the cross-power spectrum back to the spatial domain, so as to finally look for the peak location, which is no other than the translation between the two analysed images. The second process consists of refining the transformation obtained from the previous step by searching in a proximity window for a better one throughout an iterative process through pixel and two levels of subpixel accuracy, saving the best transformation as the one that provided the highest percentage in the \gls{iou} index.

In order to test the method, the transformation between 5 multispectral lenses from a MicaSense RedEdge-MX Dual camera were obtained. Just by labeling 12 images from band 5 with a high contrast allowed to obtain the transformation of \gls{bb} and mask label types with an accuracy of 97\% and 94\% and just 58 ms and 72 ms, respectively. Right after that, by using the inverse of the obtained transformations, an artificial RGB image is generated, allowing the labeling process to be performed in colored images. After it, the labels are transformed back into each lenses so as to have the labels in all 5 channels of the multispectral camera. 

Future works will consist of testing the proposed method into more multispectral cameras with different morphology, as well as testing it with all the 10 lenses that the camera used in the present paper has. Also, a RGB camera could be added in order to avoid generating fake RGB images from the multispectral lenses and accumulating a small error during the process. In a different path, a dataset of domestic waste could be created aiming to train  different \glspl{nn} and test if the extra information provided by 10 lenses and 12 bit images could help discerning better between categories when compared to the same \glspl{nn} using 8 bit RGB images.

\section*{\uppercase{Acknowledgements}}
Research work was funded by grant PID2021-122685OB-I00 funded by MICIU/AEI/10.13039/501100011033 and ERDF/EU.

\bibliographystyle{apalike}

{\small
\bibliography{paper}}

\end{document}